\pdfoutput=1
\documentclass[11pt,table,dvipsnames]{article}

\usepackage{acl}
\usepackage{times}
\usepackage{latexsym}

\usepackage[T1]{fontenc}
\usepackage[utf8]{inputenc}

\usepackage{microtype}

\usepackage{inconsolata}
\usepackage{microtype}
\usepackage{latexsym}
\usepackage{dsfont}
\usepackage{booktabs} % For formal tables
\usepackage{subfigure}
\usepackage{CJKutf8}
\usepackage{graphicx}
\usepackage{bigstrut}
\usepackage{multirow}
\usepackage{amsmath}
\usepackage{multirow, makecell, caption}
\usepackage{floatrow}
\newfloatcommand{capbtabbox}{table}[][\FBwidth]
\usepackage{xcolor}
\usepackage[normalem]{ulem}
\usepackage{amssymb}
\usepackage{amsfonts}
\usepackage{multicol}
\usepackage{tipa}
\usepackage{amsmath}
\usepackage{bm}
\usepackage[ruled,linesnumbered]{algorithm2e}
\usepackage{tikz}
\usepackage{paralist}
\usepackage{IEEEtrantools}
\usepackage[switch]{lineno}
\usepackage{enumitem}
\usepackage{multirow, makecell, caption}
\usepackage{arydshln}
\usepackage{verbatim}
\usepackage[normalem]{ulem}
\usepackage{amssymb}
\usepackage{amsfonts}
\usepackage{multicol}
\usepackage{amssymb}% http://ctan.org/pkg/amssymb
\usepackage{pifont}% http://ctan.org/pkg/pifont
\usepackage{csquotes}
\usepackage{xspace}
\usepackage{paralist}
\usepackage{mdwlist}
\usepackage{subfigure}
\usepackage{makecell}
\usepackage{array}
\usepackage{bm}
\usepackage{colortbl}
\usepackage{dsfont}
\usepackage{url}
\usepackage{booktabs} % For formal tables
\usepackage{graphicx}
\usepackage{tabularx}
\usepackage{amsmath}
\usepackage{bbm}
\usepackage{pgfplots}
\usepackage{soul} % 导入 soul 包
\usepackage{color,xcolor} % 颜色包，color 必须导入，xcolor 建议导入

\newcommand{\method}{\textsc{AnalogyKB}\xspace}
\newcommand{\probase}{Probase\xspace}
\newcommand{\wikidata}{Wikidata\xspace}
\newcommand{\conceptnet}{ConceptNet\xspace}

\newcommand{\ie}{\textit{i.e.}\xspace}
\newcommand{\eg}{\textit{e.g.}\xspace}

\newcommand{\red}[1]{{\color{red} #1}} 
\newcommand{\green}[1]{{\color{OliveGreen} #1}}

\newcommand{\mask}{\texttt{[MASK]}\xspace}

\makeatletter
\def\adl@drawiv#1#2#3{%
        \hskip.5\tabcolsep
        \xleaders#3{#2.5\@tempdimb #1{1}#2.5\@tempdimb}%
                #2\z@ plus1fil minus1fil\relax
        \hskip.5\tabcolsep}
\newcommand{\cdashlinelr}[1]{%
  \noalign{\vskip\aboverulesep
           \global\let\@dashdrawstore\adl@draw
           \global\let\adl@draw\adl@drawiv}
  \cdashline{#1}
  \noalign{\global\let\adl@draw\@dashdrawstore
           \vskip\belowrulesep}}
\makeatother

\author{Siyu Yuan\textsuperscript{\rm $\heartsuit$}\thanks{~~Equal contribution.},
 Jiangjie Chen\textsuperscript{\rm $\spadesuit$}\footnotemark[1], \\
 \bf Changzhi Sun\textsuperscript{\rm $\diamondsuit$}\thanks{~~Work done during ByteDance AI Lab.},
 Jiaqing Liang\textsuperscript{\rm $\heartsuit$}\thanks{~~Corresponding authors.},
 Yanghua Xiao\textsuperscript{\rm $\spadesuit$},
 Deqing Yang\textsuperscript{\rm $\heartsuit$}\footnotemark[3]\\
\textsuperscript{\rm $\heartsuit$}School of Data Science, Fudan University\\
\textsuperscript{\rm $\spadesuit$}Shanghai Key Laboratory of Data Science, School of Computer Science, Fudan University\\
\textsuperscript{\rm $\diamondsuit$}East China Normal University \\
\texttt{syyuan21@m.fudan.edu.cn}, \texttt{jjchen19@fudan.edu.cn},\\
\texttt{czsun.cs@gmail.com},
\texttt{\{liangjiaqing,shawyh,yangdeqing\}@fudan.edu.cn}
}

\title{\method: Unlocking Analogical Reasoning of Language Models with A Million-scale Knowledge Base}

\begin{document}

\maketitle

\begin{abstract}
Analogical reasoning is a fundamental cognitive ability of humans. 
However, current language models (LMs) still struggle to achieve human-like performance in analogical reasoning tasks due to a lack of resources for model training. 
In this work, we address this gap by proposing \method, a million-scale analogy knowledge base (KB) derived from existing knowledge graphs (KGs).
\method identifies two types of analogies from the KGs:
1) analogies of \textit{the same relations}, which can be directly extracted from the KGs, and 
2) analogies of \textit{analogous relations}, which are identified with a selection and filtering pipeline enabled by large language models (LLMs), followed by minor human efforts for data quality control.
Evaluations on a series of datasets of two analogical reasoning tasks (analogy recognition and generation) demonstrate that \method successfully enables both smaller LMs and LLMs to gain better analogical reasoning capabilities.
Resources of this paper can be found at \url{https:// github.com/siyuyuan/analogykb}.

\end{abstract}

\section{Introduction}
\label{sec:intro}

\begin{figure}[t]
    \centering
    \small
    \includegraphics[width=\linewidth]{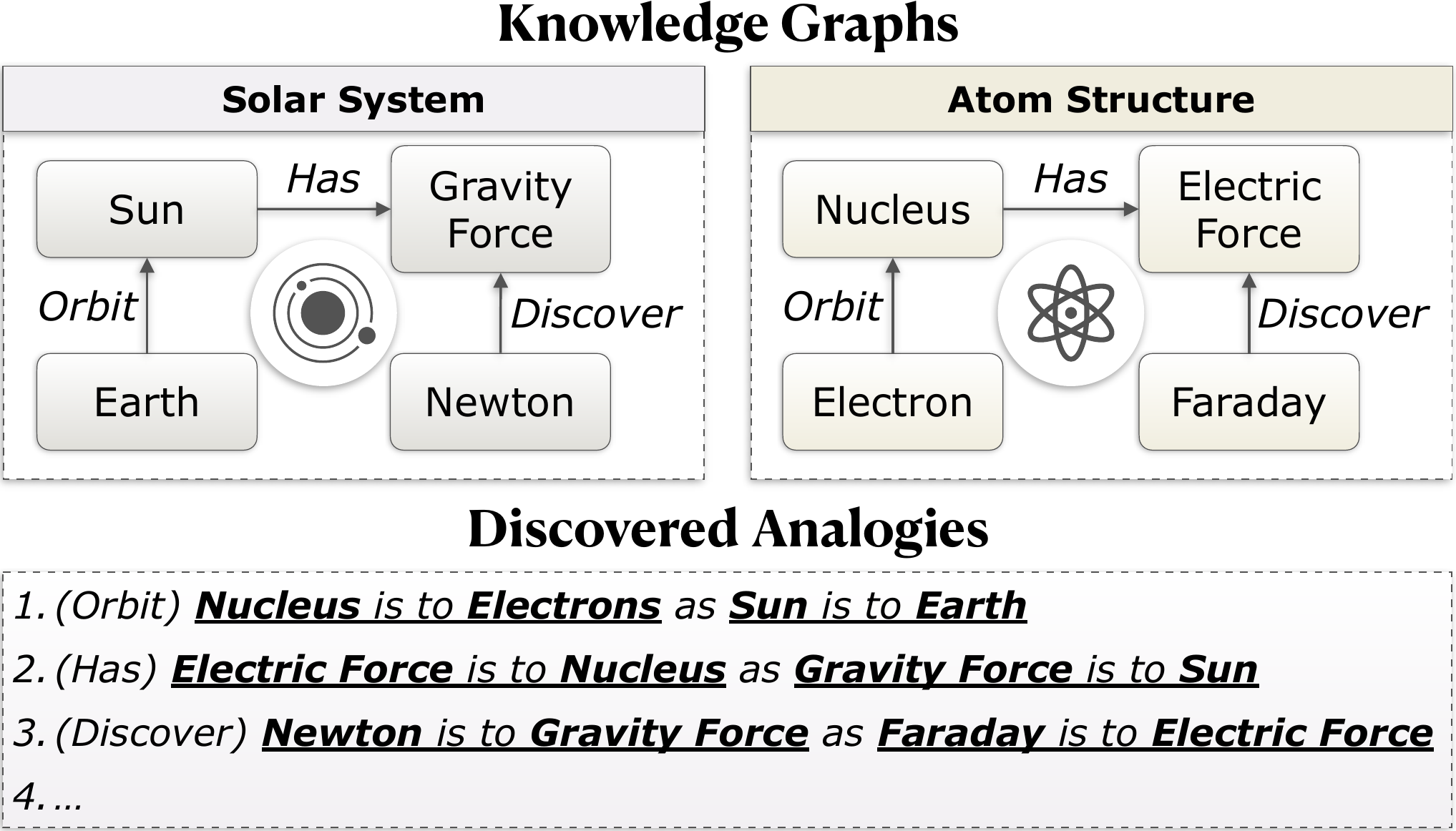}
    \caption{An example of acquiring analogies from KGs.
    Based on the relational knowledge triples from KGs, \ie, facts about the \textit{solar system} and an \textit{atom structure}, we can discover new analogies using the corresponding relations between concepts.
    }
    \label{fig:front}
\end{figure}

Making analogies requires identifying and mapping a familiar domain (\ie, source domain) to a less familiar domain (\ie, target domain)~\cite{hofstadter2013surfaces}.
As shown in Figure~\ref{fig:front}, utilizing the analogy of the solar system can facilitate comprehension of the complex structure of atoms.
Analogical reasoning is an important aspect of the cognitive intelligence of humans, allowing us to quickly adapt our knowledge to new domains~\cite{hofstadter2001analogy,10.1145/3591196.3593516}, make decisions~\cite{pmlr-v162-hansen-estruch22a}, and solve problems~\cite{yasunaga2023large}.
As a result, the topic of analogy has been drawing significant research attention in the community.

However, resources for analogical reasoning are rather limited in scale~\cite{mikolov-etal-2013-linguistic,gladkova-etal-2016-analogy,chen-etal-2022-e}, which usually consist of only hundreds or thousands of data samples.
As a result, these datasets do not support effective training of language models to gain analogical reasoning abilities. %, resulting in the fact that current state-of-the-art models still lag behind humans in ~\cite{ushio-etal-2021-bert}.
Although large language models (LLMs) can make some reasonable analogies without requiring gradient update, their performance still lies behind humans~\cite{bhavya2022analogy,jiayang2023storyanalogy}.
Therefore, larger-scale data sources are needed to facilitate the research in this area.
With richer analogies, we can train specialized analogy-making models and retrieve high-quality examples to assist LLMs.
Therefore, the research question is: \textit{How to acquire large-scale analogies at a moderate cost?}

An analogy is determined by the \textit{relational structure}~\cite{bartha2013analogy}, \eg, A:B::C:D (\ie, A is to B as C is to D), where the relation between A and B is analogous to the relation between C and D. 
The \textit{concepts} A, B, C, and D can be entities and events.
As shown in Figure~\ref{fig:front}, the ``solar system'' and an ``atom'' share a similar structure, allowing us to quickly grasp the relation between an ``electron'' and a ``nucleus'' in concepts of their source domain counterparts.
Such relational structure can be derived from the triplet knowledge, \eg (electron, orbit, nucleus) and (earth, orbit, sun), in knowledge graphs (KGs)~\cite{10.1145/2213836.2213891}.
Therefore, such structure knowledge can be utilized and reorganized to create new analogy knowledge, supporting large-scale knowledge acquisition.

In this work, we aim to build a knowledge base (KB) for storing analogies derived from existing KGs to improve analogical reasoning.
However, due to the complicated relational structures, discovering analogies from KGs is not a trivial task.
Although two pairs of concepts with the same relation can form a valid analogy (\eg, \textit{lion, isA, animal} and \textit{apple, isA, fruit}), interesting and diverse analogies are implicit in the KGs, with more complex relations.
Concepts under two distinct but similar relations in KGs can also form a reasonable analogy~\cite{hesse1959defining}.
For example, \textit{chief executive officer} and \textit{head of state} can both be abstracted into a \textit{meta relation}~\cite{hesse1959defining,gentner2017analogical}, \ie, \textit{head of organization}. 
Therefore, they are analogous relations under a meta relation.
It is important to generalize the finding of implicit analogies beyond the same relations within KGs.

We present \method, which is a large-scale analogy KB. 
We use \wikidata~\cite{10.1145/2629489} and \conceptnet~\cite{10.5555/3298023.3298212} as our seed KGs and discover two types of analogies from these KGs: analogies of 
1) \textit{same relations} and 
2) \textit{analogous relations}.
Analogies of the same relations can be directly extracted from existing KGs. 
In contrast, analogies of analogous relations are more implicit, requiring the finding of relation pairs from the KGs that can form valid analogies.
However, it is costly to manually select analogous relation pairs.
Therefore, we use InstructGPT$_\texttt{003}$~\cite{ouyang2022training}, a LLM of great capabilities in NLP tasks, for finding and deciding the analogical semantics of relations.
To eliminate the noise from the outputs of InstructGPT$_\texttt{003}$ ($\mathsection$~\ref{sec:analysis}),
we devise two filtering rules based on 1) the symmetry of analogy and 2) \textit{meta relation} summarization, which generalizes two relations into a more abstract meta relation.
Then, we \textit{manually} review the filtered results to further ensure data quality.

Our \method comprises over 1 million analogies with 943 relations, including 103 analogous relations.
Smaller LMs trained on \method gain significant improvements over the previous methods, even rivaling human performance on some analogy recognition tasks. 
Furthermore, we prove that \method can endow both smaller LMs and LLMs with satisfactory analogy-making capabilities.
Our contributions are summarized as follows:
\begin{itemize}
    \item To the best of our knowledge, we are the first to construct an analogy KB (\method) with a million scale and diverse relational structures.
    \item We propose a novel framework with LLMs to discover more interesting and implicit analogies of analogous relations;
    \item We conduct extensive experiments to evaluate the effectiveness of \method, which significantly improves the analogical reasoning performance of both smaller LMs and LLMs.
\end{itemize}

\section{Related Work}
\label{sec:related}

\paragraph{Analogy Acquisition}
Early studies mainly acquire analogy knowledge via linguists~\cite{turney2003combining,Boteanu_Chernova_2015}, which is costly and inefficient.
Recent studies consider exploiting relations in KGs to build analogies~\cite{Speer2008AnalogySpaceRT,siword,ulcar-etal-2020-multilingual}, which can be divided into two lines of work:
\begin{inparaenum}[\it 1)]
    \item \textit{Acquiring from commonsense KGs}, which leverages semantic and morphological relations from WordNet~\cite{10.1145/219717.219748}, ConceptNet~\cite{10.5555/3298023.3298212}, etc.
    However, some of these datasets are large-scale but of poor quality~\cite{li-etal-2018-analogical,li-etal-2020-ca}, while others are of high quality but limited in size~\cite{mikolov-etal-2013-linguistic,gladkova-etal-2016-analogy}.
    \item \textit{Acquiring from encyclopedia KGs}~\cite{DBLP:conf/cogsci/SiC17,zhang2022multimodal,ilievski2022does}, which utilizes the relations from DBpedia~\cite{10.1007/978-3-540-76298-0_52} and Wikidata~\cite{10.1145/2629489}, but their empirical experiments are relatively small in size.
    %, without endeavors to build large-scale analogies. 
\end{inparaenum}

\paragraph{Analogical Reasoning}
Analogical reasoning aims to identify a relational structure between two domains~\cite{bartha2013analogy,chen-etal-2022-e}.
Previous work adopts the word analogy task to investigate the analogical reasoning capability of LMs~\cite{10.5555/2999792.2999959,mikolov-etal-2013-linguistic,levy-goldberg-2014-linguistic,gladkova-etal-2016-analogy,schluter-2018-word,fournier-etal-2020-analogies,ushio-etal-2021-bert}.
Recent work demonstrates that LLMs can generate some reasonable abstract~\cite{mitchell_abstraction_2021,hu-etal-2023-context,webb2023emergent} and natural language-based analogies~\cite{bhavya2022analogy,wijesiriwardene-etal-2023-analogical,jiayang2023storyanalogy} but still lay behind humans in some cases, and smaller LMs struggle to learn analogical reasoning ability due to a lack of training data.

\paragraph{Knowledge Base Construction}

Knowledge base (KB) consists of structured knowledge to support various applications.
The approaches to constructing KBs can be divided into three categories:
\begin{inparaenum}[\it 1)]
    \item \textit{Manual construction}~\cite{10.1145/219717.219748,10.5555/3298023.3298212}, which creates the KBs with specialized knowledge written by experts, and thus is labor-intensive;
    \item \textit{Automatic construction}~\cite{10.1145/2213836.2213891,MARTINEZRODRIGUEZ2018339}, which leverages models to extract knowledge from unstructured corpora, may lead to low data quality;
    \item \textit{Semi-automatic construction}~\cite{dalvi-mishra-etal-2017-domain,10.1145/3340531.3417416}, which involves manual curation and annotation.
\end{inparaenum}
Our work is based on automatic approaches with LLMs only requiring small-scale human checking efforts.

\section{\method Construction}
\label{sec:method}

This section details the framework for building \method.
We first define the schema of \method ($\mathsection$~\ref{sec:schema}). 
Then, we collect relations with concept pairs from existing KGs ($\mathsection$~\ref{sec:Data}, Step 1)
and directly obtain analogies of the same relations from KGs ($\mathsection$~\ref{sec:Exact-relational}, Step 2). 
We propose adopting LLMs~\cite{ouyang2022training} followed by minor human efforts to acquire analogies of analogous relations ($\mathsection$~\ref{sec:Similar-relational}, Step 3).

\begin{figure}[t]
    \centering
    \includegraphics[width=\linewidth]{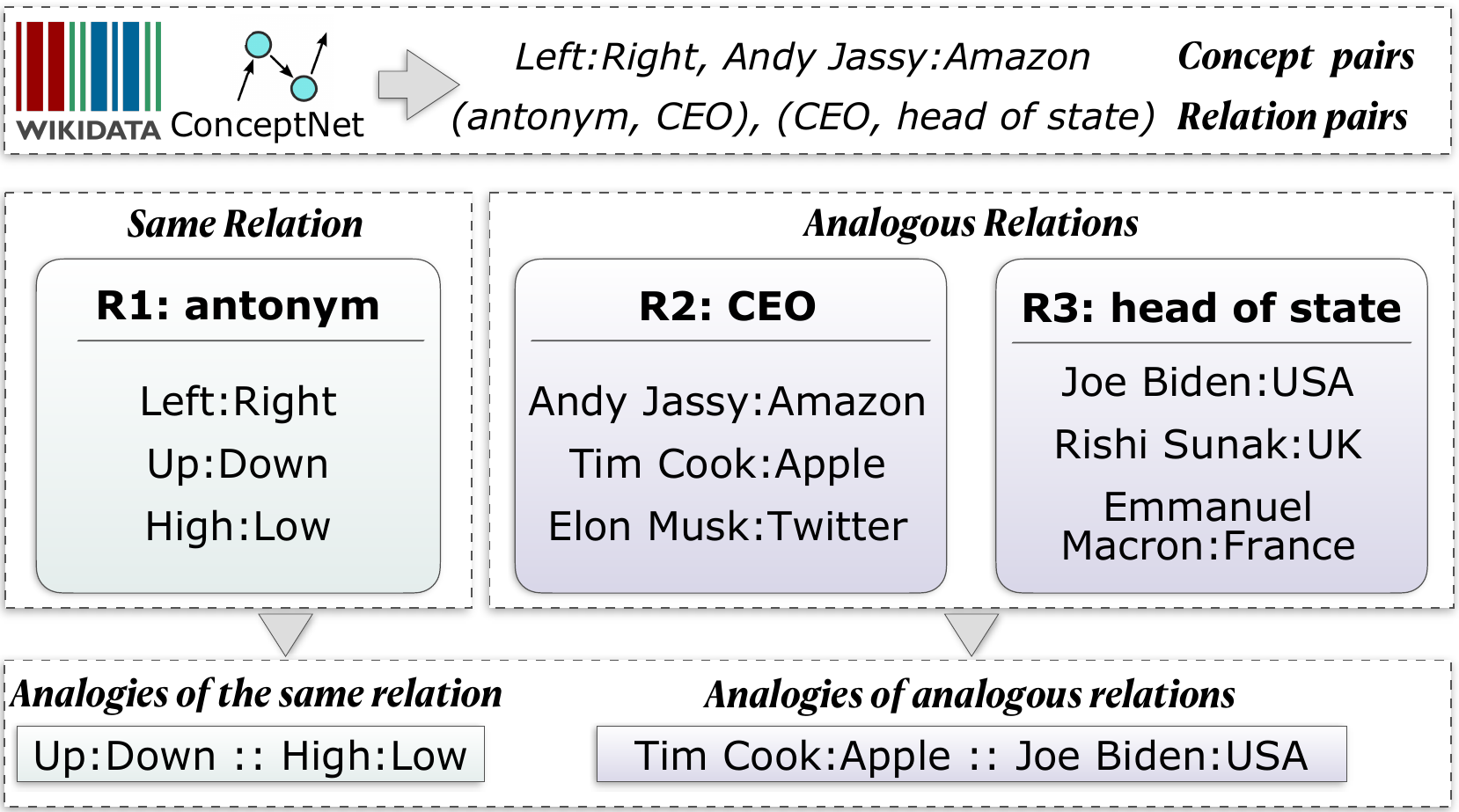}
    \caption{The relations with concept pairs are stored in \method. We define two types of analogies, \ie, analogies of \textit{the same relation} and analogies of \textit{analogous relations}, and derive them from existing KGs.}
    \label{fig:KB}
\end{figure}

\subsection{Schema for Analogies in \method}
\label{sec:schema}
This paper focuses on the analogy formed as A:B::C:D, where \textit{concepts} as A, B, C and D can be entities or events.
The \textit{concept pair} A:B is analogous to C:D based on an underlying relational structure.
Since \method is built on existing KGs, we define two types of that relational structure based on KG semantics: 
1) \textit{analogies of the same relation} and 
2) \textit{analogies of analogous relations}.
Data in \method is organized as in Figure~\ref{fig:KB}, where each relation $R$ contains subject-object \textit{concept pairs} $s:o$.
Within each relation, analogies of the same relation can be naturally formed, \eg, ``\textit{Up is to Down as High is to Low}''.
Also, the concept pairs between two relations can form analogies, as long as the \textit{relation pair} have analogous structures~\cite{hesse1959defining}.
For example, ``\textit{Tim Cook is to Apple as Joe Biden is to USA}'', where R2 (CEO) is analogous to R3 (head of state).
Therefore, \method only has to store concept pairs of each relation and analogous relation pairs, from which analogies can be easily derived.
We list the definitions of each terminology with examples in Appendix~\ref{appendix:define} for better understanding.

\subsection{Source Data Collection}\label{sec:Data}
We choose the two most-used KGs, \ie, \conceptnet and \wikidata consisting of high-quality concept pairs with relations, as our data sources.
For \conceptnet, we select the concept pairs with weights bigger than 2.0 to improve the data quality and collect 100,000 concept pairs with 27 relations.
Due to the vast amount of \wikidata, we randomly sample 5 million concepts with 813 relations from \wikidata, resulting in 20 million concept pairs.

\subsection{Acquiring Analogies of the Same Relation}
\label{sec:Exact-relational}
We can directly utilize the concept pairs in the KGs to generate analogies of the same relations.
An important perspective is that humans usually draw upon familiar domains and situations to better understand unfamiliar ones. 
To make our analogy KB more applicable to real-world scenarios, we rank the concept pairs according to their popularity scores, reflected by pageview times (in \wikidata) and concept weights (in \conceptnet).

\begin{table}[t]
\footnotesize
  \centering
    \begin{tabularx}{\linewidth}{X}
    \toprule
    \rowcolor[gray]{0.95}\multicolumn{1}{c}{\textbf{I: Analogous Relations Generation}} \\
    \midrule
    \makecell[l]{
    \color{gray}{/* \textit{I: Task prompt} */}\\
    Choose the relations from the relation candidates that can \\ form an analogy with the given relation.\\
    \color{gray}{/* \textit{Examples} */} \\
    \textbf{Given relation}: written by \\
    \textbf{Relation candidates}: [lyrics by, composed by, ...]\\ \textbf{Answer}: lyrics by, composed by, ...\\
    \color{gray}{/* \textit{Auto selection of analogical relations} */} \\
    \textbf{Given relation}: chief executive officer \\
    \textbf{Relation candidates}: [head of state, ...] \\
    \textbf{Answer}: \color[rgb]{0,0.39,0}{\textit{head of state, head of government, ...}}}\\
    
    \midrule
    
    \rowcolor[gray]{0.95}\multicolumn{1}{c}{\textbf{II: Meta Relation Summarization}} \\
    \midrule
    \makecell[l]{\color{gray}{/* \textit{Task prompt} */}\\
    Induce two relations into a higher-level relation and explain \\why they can form an analogy.\\
    \color{gray}{/* \textit{Examples} */} \\
    \makecell[l]{The relation \textbf{[lyrics by]} and the relation \textbf{[composed by]} \\ can form an analogy because both of them can be induced \\ into a relation: \textbf{[created by]}.}\\
    \makecell[l]{The relation \textbf{[written by]} and the relation \textbf{[written sys-}\\\textbf{tem]} can form an analogy because both of them can be \\induced into a relation: None.}\\
    \color{gray}{/* \textit{Auto-completion for meta relation} */} \\
     \makecell[l]{The relation \textbf{[chief executive officer]} and the relation \\ \textbf{[head of government]} can form an analogy because both \\ of them can be induced into a relation: \color[rgb]{0,0.39,0}\textit{head of organization}.}}\\
    \bottomrule
    \end{tabularx}
  \caption{Examples of prompt for InstructGPT$_\texttt{003}$ for analogous relations generation and meta relation summarization.
  {\color[rgb]{0,0.39,0}{\textit{Green texts}}} are generated by InstructGPT$_\texttt{003}$.
  }
  \label{tab:LLM_prompt}
\end{table}

\subsection{Acquiring Analogies of Analogous Relations}
\label{sec:Similar-relational}

As defined in $\mathsection$~\ref{sec:schema}, analogies of analogous relations consist of two concept pairs with analogous relations $R_1$ and $R_2$.  
However, it is difficult to automatically check whether $R_1$ and $R_2$ are analogous and manual annotation is costly.
Recently, LLMs~\cite{ouyang2022training,openai2022chatgpt} have shown their remarkable few-shot learning abilities with in-context learning.
Given a task prompt describing the task and several examples, LLMs can do the task well without training.
Therefore, we propose to exploit LLMs (\eg, InstructGPT$_\texttt{003}$) to acquire analogies of analogous relations.

\paragraph{Finding Candidate Relation Pairs}
We collect 840 relations, leading to a potential amount of $\tbinom{840}{2}$ relation pairs. 
The relations that are semantically similar to each other can form an analogy~\cite{hesse1959defining}.
For each relation, we first narrow down the candidate set from the 840 relations to the 20-most similar ones.
Specifically, we use InstructGPT embeddings (\texttt{text-embedding-ada-002}) to convert the relations into embeddings and calculate the cosine similarity between them.
By identifying the top 20 relations with the highest similarity as candidate relations for the query relation, the search space is significantly reduced for filtering analogous relations.

\paragraph{Predicting Analogous Relation Pairs}

While the search space is reduced, manual annotation remains cost-prohibitive (840 $\times$ 20).
Thus, we continue to adopt InstructGPT$_\texttt{003}$ to predict analogous relation pairs.
An example in Table~\ref{tab:LLM_prompt} (I) shows the acquisition of analogous relation pairs. 
Given examples and the query (``\textit{chief executive officer}''), InstructGPT$_\texttt{003}$ selects the relations ``\textit{head of state}'' and ``\textit{head of organization}'' from the candidates to form analogies.
Finally, InstructGPT$_\texttt{003}$ obtains 284 relation pairs.
However, we find that InstructGPT$_\texttt{003}$ struggles to filter out similar but wrong relations that cannot form analogies with queries, \eg, ``\textit{operator}'' for ``\textit{chief executive officer}'',
which requires further filtering.

\paragraph{Filtering for High-quality Relation Pairs}
In the examination process of 284 acquired relation pairs, we further implement two automatic filtering rules before conducting manual filtering to reduce human labor:
\begin{enumerate}[noitemsep]
    \item \textit{Rule 1}: if two relations can form an analogy, InstructGPT$_\texttt{003}$ should simultaneously select $R_1$ for $R_2$ and $R_2$ for $R_1$. 
    \item \textit{Rule 2}~\cite{hesse1959defining}: The second rule is using a more abstract \textit{meta relation} to decide if two relations can form an analogy. 
\end{enumerate}

The rationale behind the \textit{Rule 2} is that if two relations are analogous, then they can be generalized into a more abstract meta relation. 
For example, in Table~\ref{tab:LLM_prompt} (II), \textit{written by} and \textit{composed by} are analogous since they can be induced to a meta relation \textit{created by}.
To acquire meta relations, we prompt InstructGPT$_\texttt{003}$ with a task prompt with some examples, as shown in Table~\ref{tab:LLM_prompt} (II).
If InstructGPT$_\texttt{003}$ returns ``\textit{None}'', we discard this case.

After filtering, 103 relation pairs remain. To further improve data quality, we adopt a \textit{third} filtering by recruiting two volunteers to manually examine the remaining results, including \textbf{deleting} relation pairs that fail to form analogies or \textbf{adding} previously unchosen relation pairs that can form analogies from candidates.
Finally, we sort the concept pairs by pageview (\wikidata) and weight (\conceptnet).

\subsection{Analysis of \method}
\label{sec:analysis}

\begin{table}[t]
  \centering
  \small
    \begin{tabular}{lrrr}
    \toprule
    
    \textbf{Source} & \textbf{\# Concept Pair} & \textbf{\# Rel(s)}  & \textbf{Analogy Acc.} \\
    \midrule
    \rowcolor[gray]{0.95}\multicolumn{4}{c}{\textit{Analogies of the Same Relation}}\\
   \conceptnet & 75,019 & 27     & 98.50\%    \\
   \wikidata  & 563,024 & 813   & 98.00\%     \\
   \midrule
   \rowcolor[gray]{0.95}\multicolumn{4}{c}{\textit{Analogies of Analogous Relations}}\\
    \conceptnet & 11,829 & 5    & 95.50\%  \\
    \wikidata & 382,168 & 98     & 96.00\%      \\
    \midrule
    Total & 1,032,040 & 943     & 97.00\%     \\
    \bottomrule
    \end{tabular}%
    \caption{The statistics of \method. We report the number of concept pairs (\textbf{\# Concept Pair}) and relations (pairs if for analogous relations) (\textbf{\# Rel(s)}), manually evaluated the accuracy of randomly selected 200 analogies (\textbf{Analogy Acc.}) and the source KB (\textbf{Source}).}
  \label{tab:statistic}%
\end{table}%

\begin{table}[t]
  \centering
  \small
   \begin{tabular}{lrrc}
    \toprule
    \textbf{Data} & \textbf{\# Analogy} & \textbf{\# Rel} & \textbf{Language}\\
    \midrule
    SAT   & 374   & -     & En    \\
    Google & 550   & 15  &   En   \\
    UNIT 2 & 252   & -  &  En      \\
    UNIT 4 & 480  & -    &   En  \\
    BATS  & 1,998  & 4    &  En  \\
    E-KAR  &1251  &28   &  En \\
    E-KAR  &  1655  &  28   &  Zh  \\
    %CA-EHN & 90,505  & 763  &  Zh \\
    \cdashlinelr{1-4}
    %CA-EHN & 90k & 181k & 763    & Zh   &\xmark     &\xmark \\
    %\method & \todo    & 1002k & 918    &  En  &\cmark     &\cmark \\
    \method & $\ge$1,032,040  & 943   & En  \\
    \bottomrule
    \end{tabular}%
    \caption{Comparison between \method and previous analogy data source: numbers of analogies (\ie, A:B::C:D), number of relations and language.}
  \label{tab:compare}%
\end{table}%

\begin{figure}[t]
    \centering
    \includegraphics[width=0.9\linewidth]{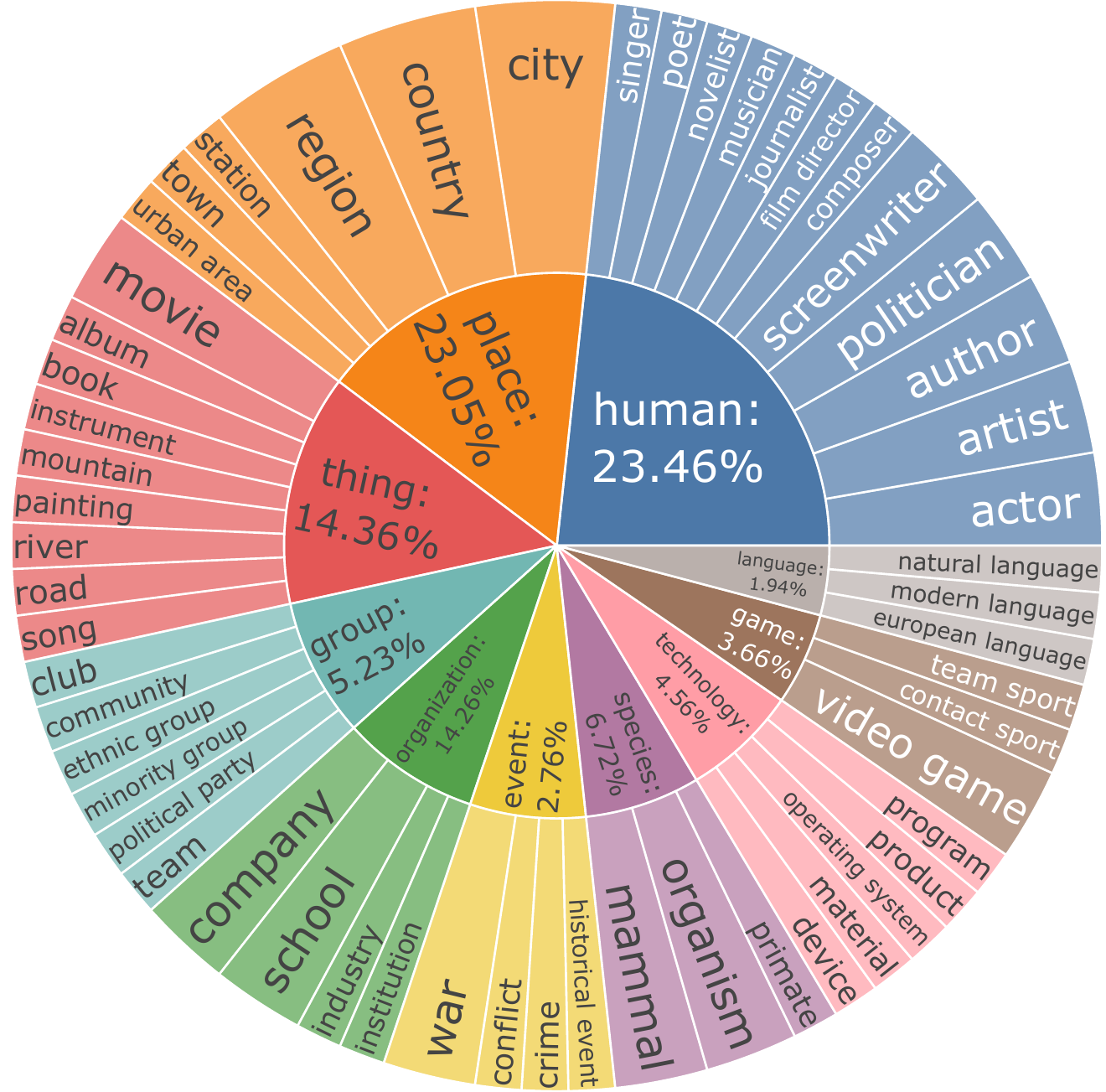}
    \caption{Distribution of concept categories in our \method. 
    }
    \label{fig:Distribution}
\end{figure}
\setlength\tabcolsep{1.8pt}
\begin{table}[t]
  \centering
  \small
    \begin{tabular}{lcc}
    \toprule
   \textbf{Method} & \textbf{\# Total} & \textbf{\# Correct} \\
    \midrule
    % GPT-3 175B & 351 & 77 \\
    % \midrule
    ChatGPT & 299 & 74\\
    InstructGPT$_\texttt{003}$ & 284 & 97\\
    %\ \ \ \ \textit{r.} \textit{SimCSE} & 343 & 9.32\\
    %\ \ \ \ \textit{r.} \textit{SBERT} & 308 & 11.36\\
    \cdashlinelr{1-3}
    \ \ \ \ + Rule 1 & 139 & 97  \\
    \ \ \ \ + Rule 1 \& Rule 2 & 103 & 97  \\
    \cdashlinelr{1-3}
    \ \ \ \ + Rule 1 \& Rule 2 \& Human & 103 & 103 \\
    \bottomrule
    \end{tabular}
    \caption{Ablated evaluation results of the analogous relation pairs.
    We record the total number of analogous relation pairs (\textbf{\# Total}) the model selects and correct ones (\textbf{\# Correct}).
    Note that ``Human'' denotes manual modifications, including adding missing relations or deleting incorrect ones, so the results are already correct (103$\rightarrow$103).
    }
  \label{tab:accuracy}%
\end{table}%
\begin{table*}[t]
  \centering
  \small
    \begin{tabular}{lccccccc}
     \toprule
    \textbf{Method} & \textbf{E-KAR}  & \textbf{BATS}  & \textbf{UNIT 2}    & \textbf{UNIT 4}    & \textbf{Google} & \textbf{SAT}   & \textbf{Mean} \\
    \midrule
    Word Embedding from RoBERTa-Large & 28.20  & 72.00  & {58.30}  & {57.40}  & 96.60  & {56.70}  & {61.53}  \\
    Word Embedding from InstructGPT &33.41	&78.30	&65.39	&62.60	&98.70	&55.38	&65.63\\
    Sentence Embedding from SentenceBERT	&25.40	&68.00	&53.40	&46.00	&90.45	&47.70	&55.16\\
    Sentence Embedding from SimCSE	&23.50	&66.54	&54.29	&50.32	&92.32	&45.10	&55.35\\
    \cdashlinelr{1-8}
    T5-Large & 40.08  & 77.37  & 34.65  & 31.25  & 75.60  & 31.45  & 48.40  \\
    BERT-Large & 36.64  & 70.10  & 32.89  & 34.49  & 90.40  & 41.30  & 50.97  \\
    
    \cdashlinelr{1-8}
    ERNIE & 40.83  & {82.54}  & 34.21  & 36.80  & 82.40  & 34.92  & 51.95  \\
    LUKE & 40.45  & {82.82}  & 34.64  & 39.12  & 88.40  & 30.26  & 52.62  \\
    \cdashlinelr{1-8}
    %InstructGPT$_\texttt{002}$ & 40.46 & 78.89    & {53.94} & 50.00  & {93.00} & 44.21 & 60.08  \\

    %InstructGPT$_\texttt{002}$ w/ CoT prompting &46.18  & 75.36  & 56.57  & 55.55  & 87.60  & 46.88  & 61.36  \\

    %InstructGPT$_\texttt{003}$ & 39.31  & 82.77  & 56.14  & 58.33  & 94.40  & 47.48  & 63.07  \\

    %InstructGPT$_\texttt{003}$ w/ CoT prompting & \underline{47.86} &	82.77 &	58.96 &	57.89 &	96.74 &	48.93 &	65.53   \\

    %ChatGPT  & 41.22  & 81.71  & 53.07  & 52.31  & 93.80  & 49.26  & 61.90  \\

    %ChatGPT w/ CoT prompting & 46.86 &	\underline{90.56} &	68.94 &	\underline{70.84} &	\underline{97.80} &	\textbf{65.78} &	\underline{73.46}   \\
    
    %\cdashlinelr{1-8}
    %RoBERTa-Large + E-KAR & {46.70}  & 79.04  & \underline{73.68}  & {61.34}  & 83.20  & 40.05  & {64.00} \\
    RoBERTa-Large & {46.70}  & 78.20  & 46.05  & 40.04  & {96.90}  & 51.60  & 59.92  \\
    \ \  + \method  & \textbf{53.43} & \underline{90.93} & \underline{87.28} & \textbf{76.15} & \underline{97.80} & \underline{59.05} & \underline{77.44} \\
    \ \  + \method (w/o check)	& 45.34	& 80.30	& 44.20	& 39.25	& 96.01	& 43.38	& 58.08 \\
    \cdashlinelr{1-8}
    DeBERTa-v3	& 47.18	& 79.54	& 50.00	& 46.99	& 96.20	& 52.26	& 62.03 \\
    \ \  + \method  & \underline{53.05}	&\textbf{92.42}	&\textbf{88.32}	&\underline{75.30}	&\textbf{98.80}	&\textbf{60.78}	&\textbf{78.11}  \\
    \ \  + \method (w/o check) &  43.89	&78.82	&45.18	&45.60	&96.00	&48.36	& 59.64 \\
    \midrule
    Human & 77.80  & 84.85  & 87.50  & 66.66  & 99.41  & 57.00  & 78.87  \\
    \bottomrule
    \end{tabular}%
    \caption{Accuracy on the analogy recognition task. We compare models and human performance on different benchmarks under different settings. The human performance values are obtained from the original papers of these analogy datasets. The best results are \textbf{bolded} and the second best ones are \underline{underlined}.}
  \label{tab:evaluation}%
\end{table*}%
As shown in Table~\ref{tab:statistic}, \method is massive, consisting of over 1 million concept pairs and 943 relations, which can form even more pairs of analogies.
Since \method provides a more comprehensive range of relations than previous datasets, it allows users to select their preferred analogies within each relation (pair). 
To evaluate the quality of \method, we randomly sample 200 analogies from each data type, \ie, two concept pairs of the same or analogous relations, in the form of A:B::C:D.
The data is annotated by two annotators with Fleiss's $\kappa=0.86$~\cite{fleiss1981measurement}.
Results show that \method is of high quality. 
Even for analogies of analogous relations, analogies are still of over 95\% accuracy.

We further compare \method with the resources related to the analogy, as reported in Table~\ref{tab:compare}.
We find that \method is much larger than previous data sources, with more analogies and relations.
To better present the fabric of \method, we present the distribution of the categories of concepts covered in \method in Figure~\ref{fig:Distribution}.
The categories are obtained from the hypernym of concepts from \probase~\cite{10.1145/2213836.2213891}. 
We find that \method exhibits high diversity.

\paragraph{Are the filtering techniques for analogous relations useful?}
We evaluate the usefulness of the filtering components, \ie, symmetry (\textit{Rule 1}) and meta relation summarization (\textit{Rule 2}), and manual correction.
We also adopt ChatGPT~\cite{openai2022chatgpt} as an ablated variant.
We record the total number of analogous relation pairs output by models (\textbf{\# Total}) and then employ annotators to report the number of correct ones out of them (\textbf{\# Correct}).
In this process, the annotators need to review these relation pairs but no need to correct them.
Each pair is examined by two annotators with Fleiss's $\kappa=0.86$.
The results in Table~\ref{tab:accuracy} show that:
\textit{1)} InstructGPT$_\texttt{003}$ is superior to ChatGPT but it still cannot filter out similar but wrong relation pairs, indicating the need for further filtering;
\textit{2)} We find the rule-based filtering technique to be rather effective, as there are not many manual corrections based on human annotations. This overcomes the labor-intensiveness of traditional KB construction methods and reveals the potential of this approach to be extended to the construction of other KBs.

\section{\method Evaluation}
\label{sec:exevalu}

\subsection{Analogy Recognition Evaluation}
\label{sec:analogy_reco}

Analogy recognition task aims to recognize the most analogous candidate to the query, formulated as multiple-choice question-answering.

\paragraph{Can models trained on \method acquire better analogy recognition abilities?}
We adopt six analogy benchmarks, \ie, E-KAR~\cite{chen-etal-2022-e}, BATS~\cite{gladkova-etal-2016-analogy}, UNIT 2 and UNIT 4~\cite{Boteanu_Chernova_2015}, Google~\cite{mikolov-etal-2013-linguistic} and SAT~\cite{turney2003combining} for evaluation. Compared to BATS and Google, E-KAR, UNIT 2, UNIT 4, and SAT contain more abstract and complex analogies and thus more difficult for humans.

For the backbone model, we use the RoBERTa-Large~\cite{liu2019roberta} and randomly sample 10,000 data points from \method to train the model in a multiple-choice question-answer format. 
We first train the model on the data from \method and then further fine-tune it on 
benchmarks.\footnote{Detailed information on the benchmarks is shown in Appendix~\ref{appendix:Benchmark} and the construction of \method sample data is shown in Appendix~\ref{appendix:ART_dataset}.}
For baselines, we adopt pre-trained word embeddings~\cite{ushio-etal-2021-bert,ouyang2022training}, pre-trained sentence embeddings~\cite{reimers-gurevych-2019-sentence,gao-etal-2021-simcse}, pre-trained language models~\cite{10.5555/3455716.3455856,devlin-etal-2019-bert,liu2019roberta,he2023debertav}.
To rule out the confounder in \method, we also add knowledge-enhanced models, ERNIE~\cite{zhang-etal-2019-ernie} and LUKE~\cite{yamada-etal-2020-luke} which contain the relational knowledge between entities.
Moreover, we also randomly sampled 10,000 data points from the \method without checking and filtering, \ie, + \method (w/o check), to prove the necessity of filtering. 
After human examination, nearly about 63\% of data points do not form analogies.
Previous benchmarks, except E-KAR, do not have a training set. Thus, we fine-tune LMs on their small development set.\footnote{Details about the baselines and training process are shown in Appendix~\ref{appendix:models} and \ref{appendix:ART_training}. The statistical significant test is shown in Appendix~\ref{sec:significant}}

The results presented in Table~\ref{tab:evaluation} show that:
\begin{inparaenum}[\it 1)]
    \item The performance of sentence embeddings is inferior to word embeddings. The rationale is that such word analogy is based on relational rather than semantic similarity between two sentences. Therefore, taking the difference between two word embeddings is a more reasonable yet still problematic approach for finding word analogies.
    \item Incorporating entity knowledge cannot improve model performance on analogy recognition;
    \item The training data without checking brings noise and even degrades model performance, further emphasizing the importance of high-quality data in \method.
    \item Training models on \method can significantly improve the model performance on analogy recognition by a large margin.
\end{inparaenum}

\begin{figure}[t]
    \centering
    \small
    \includegraphics[width=\linewidth]{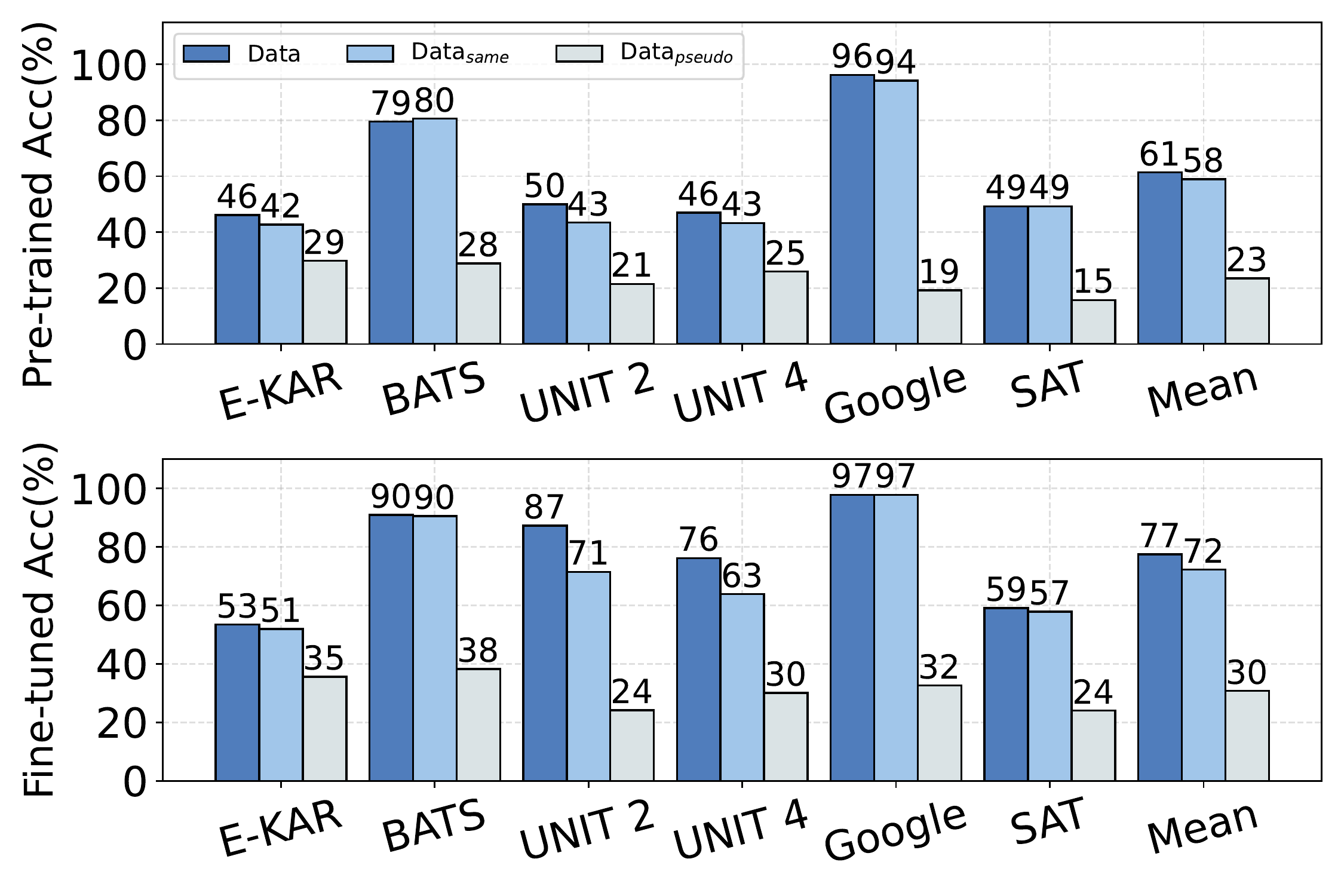}
    \caption{The accuracy of RoBERTa-Large trained on different data subsets on the analogy recognition task. 
    \textbf{Data} denotes the dataset sampled directly from \method, \textbf{Data$_\texttt{same}$} denotes the dataset that only has same-relation analogies, and \textbf{Data$_\texttt{pseudo}$} denotes the dataset with concept pairs that do not form analogies. All the datasets have the same size.
    }
    \label{fig:ablation}
\end{figure}

\paragraph{How much do analogies of analogous relations in \method contribute to performance?}

We create two ablated variants from \method to train the models:
\begin{inparaenum}[\it 1)]
    \item \textit{Analogies of the same relations}, denoted as Data$_\texttt{same}$: we randomly sampled 10,000 data of the same relations as an ablated variant. 
    \item \textit{Pseudo analogies}, denoted as Data$_\texttt{pseudo}$: 
    for each data point, we randomly sample 5 concept pairs from the \method and choose one as the query, one as the answer, and the remaining three as distractions. 
    This makes sure that \method indeed imposes analogical reasoning ability on the model rather than simply data augmentation.
\end{inparaenum}
We adopt two settings: only train RoBERTa-Large on 10,000 data (\ie, \textbf{Pre-trained}) and first train RoBERTa-Large on 10,000 data and then fine-tune it on the specific benchmarks (\ie, \textbf{Fine-tuned}).

The results in Figure~\ref{fig:ablation} show that:
\begin{inparaenum}[\it 1)]
    \item Analogies of analogous relations in \method are rather important for models to comprehend analogies with more abstract and complex relations.
    \item Training models on randomly constructed analogy-style data even drags down model performance, further emphasizing the importance of \method.\footnote{We also compare the data from different KB sources, which is shown in Appendix~\ref{appendix:different_KB}.}
\end{inparaenum}

\begin{figure}[t]
    \centering
	\subfigure[Different data sizes.] {
    \label{fig:abl_size_data} 
		\pgfplotsset{width=0.345\linewidth,height=0.33\linewidth,compat=1.15,scale only axis}
% \begin{figure}[th]
% \begin{centering}
\footnotesize
% \centering
\begin{tikzpicture}
\begin{axis}[
    % axis y line*=left,
    xlabel={Data Size ($\times$1k)},
    xmin=0, xmax=150,
    ymin=20, ymax=55,
    legend pos=south east,
    % nodes near coords, 
    % nodes near coords align={vertical},
    ymajorgrids=true,
    xmajorgrids=true,
    grid style=dashed,
    x label style={at={(axis description cs:0.5,-0.125)},anchor=north},%,font=\tiny},
    % y label style={at={(axis description cs:-0.125,0.5)},anchor=south},%,font=\tiny},
    legend style={nodes={scale=0.7, transform shape},font=\footnotesize},
    xmode=log
]
\addplot[
    color=NavyBlue,
    mark=o,
    mark size=1.5pt,
    error bars/.cd,
    y dir=both, y explicit
    ]
    coordinates {
    (0, 34.35)% +- (0.0, 0.0)
    (1, 37.78)% +- (0.0, 0.0)
    (5, 40.46)% +- (0.0, 0.0)
    (10, 44.06)% +- (0.0, 0.0)
    (50, 46.18)% +- (0.0, 0.0)
    (100, 46.61)
    };
    \addlegendentry{AKB}
\addplot[
    color=Maroon,
    mark=triangle,
    mark size=2pt,
    error bars/.cd,
    y dir=both, y explicit
    ]
    coordinates {
    (0, 40.08)% +- (0.0, 0.0)
    (1, 42.37)% +- (0.0, 0.0)
    (5, 43.64)% +- (0.0, 0.0)
    (10, 45.04)% +- (0.0, 0.0)
    (50, 46.95)% +- (0.0, 0.0)
    (100, 47.45)
    };
    % \label{plot:ratio_tn_CG}
    \addlegendentry{AKB + E-KAR}
\end{axis}
\end{tikzpicture}}
	\subfigure[Different model sizes.] { 
	\label{fig:abl_size_model} 		\pgfplotsset{width=0.345\linewidth,height=0.33\linewidth,compat=1.15,scale only axis}
% \begin{figure}[th]
% \begin{centering}
\footnotesize
% \centering
\begin{tikzpicture}
\begin{axis}[
    % axis y line*=left,
    xlabel={Model Size (M)},
    ymin=20, ymax=55,
    % xtick={20,40,60,80},
    % xmode=log
    legend pos=south east,
    % nodes near coords, 
    % nodes near coords align={vertical},
    ymajorgrids=true,
    xmajorgrids=true,
    grid style=dashed,
    x label style={at={(axis description cs:0.5,-0.125)},anchor=north},%,font=\tiny},
    % y label style={at={(axis description cs:-0.125,0.5)},anchor=south},%,font=\tiny},
    legend style={nodes={scale=0.7, transform shape},font=\footnotesize},
    xmode=log
]
\addplot[
    color=NavyBlue,
    mark=o,
    mark size=1.5pt,
    error bars/.cd,
    y dir=both, y explicit
    ]
    coordinates {
    (60, 29.01)% +- (0.0, 0.0)
    (220, 42.75)% +- (0.0, 0.0)
    (770, 44.06)% +- (0.0, 0.0)
    (3072, 50.76)% +- (0.0, 0.0)
    };
    \addlegendentry{AKB}
\addplot[
    color=Maroon,
    mark=triangle,
    mark size=2pt,
    error bars/.cd,
    y dir=both, y explicit,
    ]
    coordinates {
    (60, 32.44)% +- (0.0, 0.0)
    (220, 43.13)% +- (0.0, 0.0)
    (770, 45.04)% +- (0.0, 0.0)
    (3072, 51.53)% +- (0.0, 0.0)
    };
    % \label{plot:ratio_tn_CG}
    \addlegendentry{AKB + E-KAR}
\end{axis}
\end{tikzpicture}}
    \caption{Performance change (Accuracy \%) for T5 on E-KAR test set with increasing training data (1K, 5K, 10K, 50K, 100K) from \method and model size (60M, 220M, 770M, 3B).
    T5 is either trained on \method (AKB) or both \method and E-KAR (AKB + E-KAR).
    }
    \label{fig:size}
\end{figure}
\paragraph{How do data sizes and model sizes affect performance?}

We use T5-Large as the base model to examine the effects of training data size on model performance.
We first train the model on data from \method, and fine-tune it on E-KAR.
As illustrated in Figure~\ref{fig:abl_size_data}, increasing the amount of training data from \method improves model performance. 
Figure~\ref{fig:abl_size_model} shows the results of different-sized T5 models on 10,000 data points from \method.
We find that the larger models get less of a performance gain from E-KAR, indicating that they learn more from \method and can better generalize to E-KAR.

\setlength\tabcolsep{4pt}

\begin{table}[t]
  \centering
  \small
    \begin{tabular}{lccc}
    \toprule
    \textbf{Model}  & \textbf{E-KAR} & \textbf{UNIT 4} & \textbf{SAT} \\
    \midrule
    {vanilla T5}          & 13.00 & 17.00 & 8.00 \\
    {AnalogyT5$_\texttt{same}$}           & \underline{42.00} & \underline{63.00} & \underline{37.00} \\
    {AnalogyT5}           & \textbf{57.00} & \textbf{80.00} & \textbf{64.00} \\
    \midrule
    %{InstructGPT$_\texttt{002}$}& 50.00 & 63.00 & 43.00 \\
    %\ \ + {Human} & 55.00 & 72.00 & 57.00 \\
    %\ \ + {\textsc{AnalogyKB}$_\texttt{same}$} & 58.00 & 72.00 & 66.00 \\
    %\ \ + {\method} & 68.00 & 76.00 & 74.00 \\
    %\cdashlinelr{1-4}
    {InstructGPT$_\texttt{003}$}& 61.00 & 70.00 & 60.00 \\
    \ \ + {Human} & 68.00 & 76.00 & 74.00 \\
    \ \ + {\textsc{AnalogyKB}$_\texttt{same}$} & 64.00 & 77.00 & 77.00 \\
    \ \ + {\method} & \textbf{75.00} & 80.00 & \underline{85.00} \\
    \cdashlinelr{1-4}
    {ChatGPT}       & 58.00 & 76.00 & 78.00 \\
    \ \ + {Human} & 64.00 & \underline{81.00} & 80.00 \\
    \ \ + {\textsc{AnalogyKB}$_\texttt{same}$} & 64.00 & 80.00 & 81.00 \\
    \ \ + {\method} & \underline{69.00} & \textbf{92.00} & \textbf{91.00} \\
    \bottomrule
    \end{tabular}
    \caption{
    Accuracy on analogy generation. 
    For LLMs, we compare LLMs with 0-shot and human-written examples (+ Human) vs. \method-retrieved examples (+ \method).
    %0-shot learning and with 8 examples written by humans (+ Human) to that with 8 examples retrieved from \method (+ \method).
    For smaller LMs, we compare AnalogyT5 with vanilla T5.
    AnalogyT5$_\texttt{same}$ and \textsc{AnalogyKB}$_\texttt{same}$ are the ablation variants with analogies of the same relations from \method.
    %The best results are \textbf{bolded} and the second best ones are \underline{underlined}.
    }
  \label{tab:retrieve}%
\end{table}%

\begin{table}[t]
  \centering
  \small
    \begin{tabular}{lrrr}
     \toprule
    \textbf{Model} & \textbf{Acc.} & \textbf{MRR} & \textbf{Rec@5}\\
    \midrule
    %Glove & 1.80 & 2.20  & 2.40 \\
    %\midrule
   GPT-2  & 2.00 & 5.70 & 10.70  \\
    \ \ + BATS & 1.10 & 2.20 & 3.60 \\
    \ \ + E-KAR	&2.39	&5.44	&10.46 \\
    \ \ + \textsc{AnalogyKB}$_\texttt{same}$ & 4.00 &	5.49 	&11.45 \\
    \ \ + \method & 5.12 &  6.41 & 12.77 \\
    \cdashlinelr{1-4}
    BERT & 0.90 & 4.40 & 8.00  \\
    \ \ + BATS & 0.40 & 1.90 & 3.10  \\
    \ \ + E-KAR	&1.50	&4.32	&7.92 \\
    \ \ + \textsc{AnalogyKB}$_\texttt{same}$ & 4.01  & 7.44  & 10.89  \\
    \ \ + \method & 6.24 & {10.36} & {14.07}  \\
\midrule
   %  InstructGPT$_\texttt{002}$ & 3.00 & 8.11 & 22.39  \\
   %\ \ + BATS & 4.33 & 10.52 & 27.49   \\
   %\ \ + \textsc{AnalogyKB}$_\texttt{same}$ & 5.23 &	10.21 	& 23.43 \\
   %\ \ + \method & \underline{12.80} & 17.97 & \underline{35.33}   \\
   %\cdashlinelr{1-4}
   InstructGPT$_\texttt{003}$ & 3.32 & 15.75 & 34.58  \\
   \ \ + BATS & 5.07 & {21.40} & 32.37    \\
   \ \ + E-KAR	&9.12&	25.00	&36.27 \\
   \ \ + \textsc{AnalogyKB}$_\texttt{same}$ & 6.91 &	\underline{25.32} 	& 33.42 \\
   \ \ + \method & \textbf{15.30} & \textbf{32.80} & \textbf{38.46}   \\
    \bottomrule
    \end{tabular}%
    \caption{
    Analogy generation results on SCAN.
    For LLMs, we compare LLMs with 0-shot and examples retrieved from BATS (+ BATS) and E-KAR (+ E-KAR) vs. retrieved from \method (+ \method).
    For smaller LMs, we pre-train the models on BATS (+ BATS) or E-KAR (+ E-KAR) or data sampled from \method (+ \method). 
    %\textsc{AnalogyKB}$_\texttt{same}$ is the ablation variants with analogies of the same relations from \method.
    %The best results are \textbf{bolded} and the second best ones are \underline{underlined}.
    }
  \label{tab:scan}%
\end{table}%

\subsection{Analogy Generation Evaluation}
\label{sec:AGE}

This task can be formulated as a text generation task: completing the \textit{D} given \textit{A}, \textit{B}, \textit{C} to form a plausible analogy \textit{A is to B as C is to D}.
Analogy generation is of more practical use, since the generation of familiar analogies could be helpful to comprehend the source problem.

\paragraph{Does \method support analogy generation?}
To answer this question, we investigate two settings: For smaller LMs, we randomly sample 1 million data points from \method. Then we fine-tune T5-Large on \method (named AnalogyT5) to compare vanilla T5.
For LLMs, we convert the query and analogies from \method into InstructGPT embeddings, retrieve the top-8 most similar analogies based on cosine embedding similarity, and use them as examples in the prompt.
We test models on 100 test data sampled from three challenging benchmarks, which are not found in the training set.\footnote{Detailed information on the training process and the results on six benchmarks are shown in Appendix~\ref{appendix:Retrieval_training} and \ref{sec:six_benchmarks}. We also conduct the impact of data and model sizes and case studies for further analysis in Appendix~\ref{sec:impact} and \ref{sec:case_study}.} 
Each generation is evaluated by three annotators with Fleiss's $\kappa=0.93$.
The results in Table~\ref{tab:retrieve} show that, in both pre-training and in-context learning, \method enables better analogy generation, and the analogies of analogous relations prove significantly valuable to the performance of models.

\paragraph{Does \method help LMs generalize to out-of-domain analogies?}
%Despite its high coverage of common concepts ($\mathsection$~\ref{sec:analysis}), \method contains few analogies related to metaphor and science which are abstract and complex, posing a challenge even for humans~\cite{czinczoll2022scientific}.
Despite its high coverage of common concepts ($\mathsection$~\ref{sec:analysis}), \method contains few analogies related to metaphor and science which are not common in the KGs and thus out-of-domain.
%These types of data are not present in existing KGs and are hard to extract from corpora.
To examine whether \method can generalize the ability of LMs to reason about these analogies, we test AnalogyT5 on the SCAN dataset~\cite{czinczoll2022scientific}, which has 449 analogies of metaphor and science domains. 
For smaller LMs, we follow the original experimental setup and compare the models trained on \method (see Appendix~\ref{appendix:DAT} for details). 
For LLMs, we retrieve the top-8 most similar analogies from \method as examples, in contrast to zero-shot settings, retrieving from BATS and E-KAR.
The results shown in Table~\ref{tab:scan} reveal that
\begin{inparaenum}[\it 1)]
    \item For smaller LMs, training on BATS even worsens performance on SCAN. 
    However, training on E-KAR with complex analogies can indeed improve the model performance on SCAN.  % indicates that previous datasets are insufficient for LMs to make complex analogies.
    \item Compared to E-KAR, \method can further help both LLMs and smaller models generalize to out-of-domain analogies.
\end{inparaenum}

\begin{figure}[t]
    \centering
    \small
    \includegraphics[width=\linewidth]{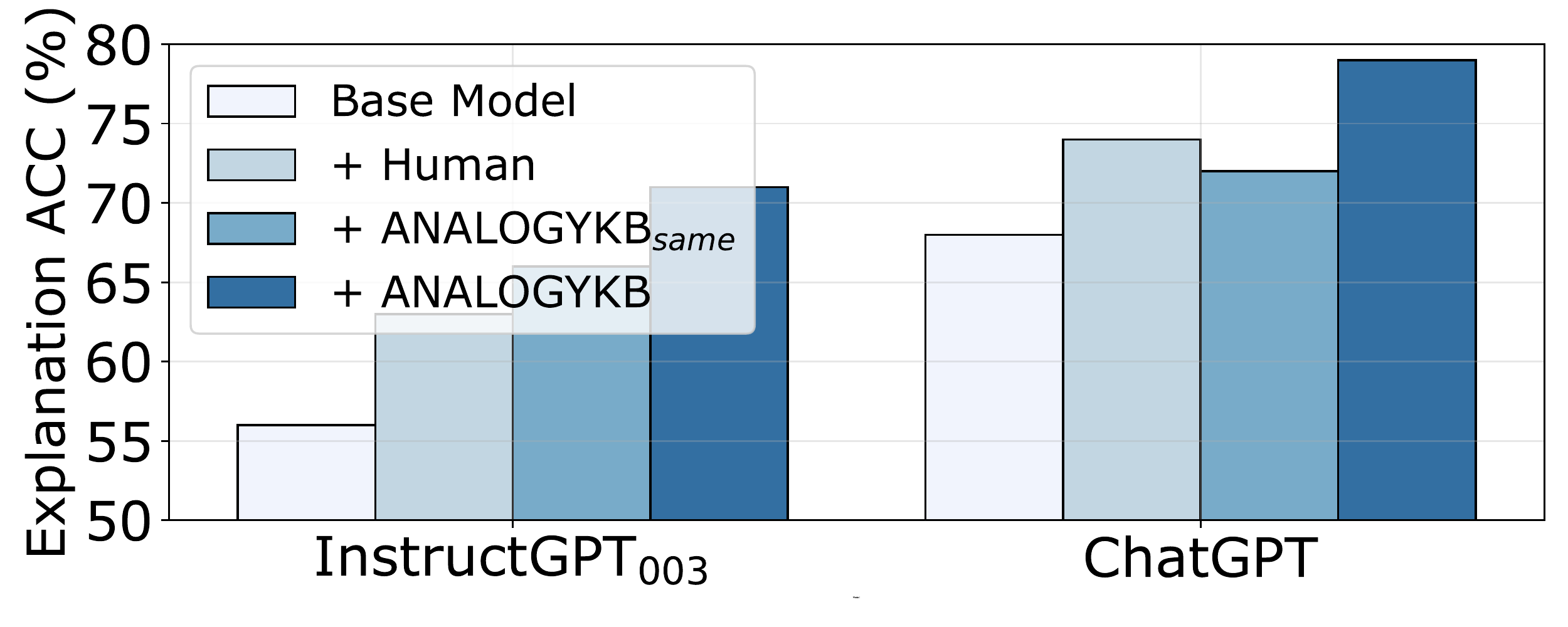}
    \caption{The accuracy of LLMs on the analogy explanation task. We compare LLMs with 0-shot (Base Model) and human-written examples (+ Human) vs. \method-retrieved examples (+ \method).
    }
    \label{fig:Explanation}
\end{figure}

\paragraph{Can \method better support analogy explanation for LLMs?}
%We have proven that \method can help LMs in analogy generation. 
%However, a more intriguing question arises: does \method support analogy explanation? 
Analogy explanation needs LLMs to provide a reasonable explanation for a given analogy, which more closely simulates the process of human reasoning and knowledge explanation.
%The analogy explanation can utilize \method in a more effective way for future research that, 
%To , we first convert the query and analogies from \method into embeddings and retrieve the top-8 most similar analogies based on cosine embedding similarity.
In this setting, we first retrieve top-8 most similar analogies based on cosine embedding similarity.
Then, we ask GPT-4 to generate explanations for the analogies given relations and use them as examples in the prompt. 
We test InstructGPT$_\texttt{003}$ and ChatGPT on 100 data samples from E-KAR, and employ two annotators to judge whether the explanations are correct with Fleiss's $\kappa= 0.97$).
The results in Figure~\ref{fig:Explanation} are consistent with Table~\ref{tab:retrieve}, demonstrating that \method can facilitate better analogy explanation for LLMs, and the analogies of analogous relations are significantly valuable for performance.

\section{Conclusion}
\label{sec:conclusion}

In this paper, we introduce \method, a million-scale analogy KB to improve model performance in analogical reasoning tasks. 
We identify two types of analogies in existing KGs, \ie, analogies of the same and analogous relations, and utilize LLMs with minor human examinations to find them.
\method demonstrates its great value in assisting both smaller LMs and LLMs with the resolution of analogy recognition and generation tasks, especially with analogies of analogous relations in \method.

\section*{Limitations}
\label{sec:limitation}

First, this paper only considers analogies involving one or two relations and primarily concentrates on analogies in the form of ``A is to B as C is to D''.
However, analogies may involve the combination of multiple relations of multiple entities or even events. 
For example, an engineer can learn the eye cross-section by taking the analogy of the camera structure. 
Here, the analogy involves multiple entities and relations in the two systems (camera and eye): Aperture should be analogous to pupil since both are channels for light to enter and black paint should be analogous to choroid since both absorb light to prevent it from bouncing and reflecting.

Second, our \method is constructed using data from \wikidata and \conceptnet, which do not include analogies in other domains such as the scientific domain. 
For example, it would be challenging for LMs trained on \method to reason about an analogy such as Protein synthesis in a cell is like a factory assembly line as it would require a deep understanding of biological and industrial processes, which is not well-covered in our data sources.
Also, \method is stored in the form of tuples, but in practice, some analogy situations may not be easily converted to this format. 
Future research should address how to bridge this gap.

Due to the limited computational resources, we only use a subset of \method. Assuming unlimited computational resources, the far-stretching goal of this project is to enable the discovery of new, better analogies for applications such as explanation (e.g., science popularization), text polishing, and case-based reasoning. So, with the full scale of the data, we can train a specialized open-source large language model (e.g., Llama 2) in such related tasks with data from \method so that these models can discover novel analogies and understand new concepts and knowledge with analogical reasoning ability.

\section*{Ethics Statement}
\label{sec:Ethics}
We hereby acknowledge that all authors of this work are aware of the provided ACL Code of Ethics and honor the code of conduct.

\paragraph{Use of Human Annotations}
The annotations of relation pairs in \method are implemented by annotators recruited by our institution.
The construction team remains anonymous to the authors, and the annotation quality is ensured by using a double-check strategy as described in Section~\ref{sec:method}.
We ensure that the privacy rights of all annotators are respected throughout the annotation process. All annotators are compensated above the local minimum wage and consent to the use of \method for research purposes, as described in our paper. The annotation details are shown in Appendix~\ref{appendix:crowdsourcing}.
\paragraph{Risks}
The database is sourced from publicly available sources, \wikidata and \conceptnet. 
However, we cannot guarantee that it is free of socially harmful or toxic language. 
Additionally, analogy evaluation relies on commonsense, and different individuals with diverse backgrounds may have varying perspectives.

\section*{Acknowledgement}
We thank the anonymous reviewers for their valuable comments.
This work is supported by the Chinese NSF Major Research Plan (No.92270121), Shanghai Science and Technology Innovation Action Plan (No.21511100401) and the Science and Technology Commission of Shanghai Municipality Grant (No. 22511105902).

\bibliography{anthology,custom}

\begin{thebibliography}{52}
\expandafter\ifx\csname natexlab\endcsname\relax\def\natexlab#1{#1}\fi

\bibitem[{Allen and Hospedales(2019)}]{siword}
Carl Allen and Timothy Hospedales. 2019.
\newblock Analogies explained: Towards understanding word embeddings.
\newblock In \emph{International Conference on Machine Learning}, pages
  223--231. PMLR.

\bibitem[{Auer et~al.(2007)Auer, Bizer, Kobilarov, Lehmann, Cyganiak, and
  Ives}]{10.1007/978-3-540-76298-0_52}
S{\"o}ren Auer, Christian Bizer, Georgi Kobilarov, Jens Lehmann, Richard
  Cyganiak, and Zachary Ives. 2007.
\newblock Dbpedia: A nucleus for a web of open data.
\newblock In \emph{The Semantic Web}, pages 722--735, Berlin, Heidelberg.
  Springer Berlin Heidelberg.

\bibitem[{Bartha(2013)}]{bartha2013analogy}
Paul Bartha. 2013.
\newblock Analogy and analogical reasoning.

\bibitem[{Bhavya et~al.(2022)Bhavya, Xiong, and Zhai}]{bhavya2022analogy}
Bhavya Bhavya, Jinjun Xiong, and Chengxiang Zhai. 2022.
\newblock Analogy generation by prompting large language models: A case study
  of instructgpt.
\newblock \emph{arXiv preprint arXiv:2210.04186}.

\bibitem[{Boteanu and Chernova(2015)}]{Boteanu_Chernova_2015}
Adrian Boteanu and Sonia Chernova. 2015.
\newblock \href {https://doi.org/10.1609/aaai.v29i1.9400} {Solving and
  explaining analogy questions using semantic networks}.
\newblock \emph{Proceedings of the AAAI Conference on Artificial Intelligence},
  29(1).

\bibitem[{Chen et~al.(2022)Chen, Xu, Fu, Shi, Li, Zhang, Sun, Li, Xiao, and
  Zhou}]{chen-etal-2022-e}
Jiangjie Chen, Rui Xu, Ziquan Fu, Wei Shi, Zhongqiao Li, Xinbo Zhang, Changzhi
  Sun, Lei Li, Yanghua Xiao, and Hao Zhou. 2022.
\newblock \href {https://doi.org/10.18653/v1/2022.findings-acl.311} {{E}-{KAR}:
  A benchmark for rationalizing natural language analogical reasoning}.
\newblock In \emph{Findings of the Association for Computational Linguistics:
  ACL 2022}, pages 3941--3955, Dublin, Ireland. Association for Computational
  Linguistics.

\bibitem[{Czinczoll et~al.(2022)Czinczoll, Yannakoudakis, Mishra, and
  Shutova}]{czinczoll2022scientific}
Tamara Czinczoll, Helen Yannakoudakis, Pushkar Mishra, and Ekaterina Shutova.
  2022.
\newblock Scientific and creative analogies in pretrained language models.
\newblock \emph{arXiv preprint arXiv:2211.15268}.

\bibitem[{Dalvi~Mishra et~al.(2017)Dalvi~Mishra, Tandon, and
  Clark}]{dalvi-mishra-etal-2017-domain}
Bhavana Dalvi~Mishra, Niket Tandon, and Peter Clark. 2017.
\newblock \href {https://doi.org/10.1162/tacl_a_00058} {Domain-targeted, high
  precision knowledge extraction}.
\newblock \emph{Transactions of the Association for Computational Linguistics},
  5:233--246.

\bibitem[{Devlin et~al.(2019)Devlin, Chang, Lee, and
  Toutanova}]{devlin-etal-2019-bert}
Jacob Devlin, Ming-Wei Chang, Kenton Lee, and Kristina Toutanova. 2019.
\newblock \href {https://doi.org/10.18653/v1/N19-1423} {{BERT}: Pre-training of
  deep bidirectional transformers for language understanding}.
\newblock In \emph{Proceedings of the 2019 Conference of the North {A}merican
  Chapter of the Association for Computational Linguistics: Human Language
  Technologies, Volume 1 (Long and Short Papers)}, pages 4171--4186,
  Minneapolis, Minnesota. Association for Computational Linguistics.

\bibitem[{Ding et~al.(2023)Ding, Srinivasan, Macneil, and
  Chan}]{10.1145/3591196.3593516}
Zijian Ding, Arvind Srinivasan, Stephen Macneil, and Joel Chan. 2023.
\newblock \href {https://doi.org/10.1145/3591196.3593516} {Fluid transformers
  and creative analogies: Exploring large language models’ capacity for
  augmenting cross-domain analogical creativity}.
\newblock In \emph{Proceedings of the 15th Conference on Creativity and
  Cognition}, C\&C '23, page 489–505, New York, NY, USA. Association for
  Computing Machinery.

\bibitem[{Fleiss et~al.(1981)Fleiss, Levin, Paik
  et~al.}]{fleiss1981measurement}
Joseph~L Fleiss, Bruce Levin, Myunghee~Cho Paik, et~al. 1981.
\newblock The measurement of interrater agreement.
\newblock \emph{Statistical methods for rates and proportions},
  2(212-236):22--23.

\bibitem[{Fournier et~al.(2020)Fournier, Dupoux, and
  Dunbar}]{fournier-etal-2020-analogies}
Louis Fournier, Emmanuel Dupoux, and Ewan Dunbar. 2020.
\newblock \href {https://doi.org/10.18653/v1/2020.conll-1.29} {Analogies minus
  analogy test: measuring regularities in word embeddings}.
\newblock In \emph{Proceedings of the 24th Conference on Computational Natural
  Language Learning}, pages 365--375, Online. Association for Computational
  Linguistics.

\bibitem[{Gao et~al.(2021)Gao, Yao, and Chen}]{gao-etal-2021-simcse}
Tianyu Gao, Xingcheng Yao, and Danqi Chen. 2021.
\newblock \href {https://doi.org/10.18653/v1/2021.emnlp-main.552} {{S}im{CSE}:
  Simple contrastive learning of sentence embeddings}.
\newblock In \emph{Proceedings of the 2021 Conference on Empirical Methods in
  Natural Language Processing}, pages 6894--6910, Online and Punta Cana,
  Dominican Republic. Association for Computational Linguistics.

\bibitem[{Gentner and Maravilla(2017)}]{gentner2017analogical}
Dedre Gentner and Francisco Maravilla. 2017.
\newblock Analogical reasoning.
\newblock In \emph{The Routledge International Handbook of Thinking and
  Reasoning}, pages 186--203. Routledge.

\bibitem[{Gladkova et~al.(2016)Gladkova, Drozd, and
  Matsuoka}]{gladkova-etal-2016-analogy}
Anna Gladkova, Aleksandr Drozd, and Satoshi Matsuoka. 2016.
\newblock \href {https://doi.org/10.18653/v1/N16-2002} {Analogy-based detection
  of morphological and semantic relations with word embeddings: what works and
  what doesn{'}t.}
\newblock In \emph{Proceedings of the {NAACL} Student Research Workshop}, pages
  8--15, San Diego, California. Association for Computational Linguistics.

\bibitem[{Hansen-Estruch et~al.(2022)Hansen-Estruch, Zhang, Nair, Yin, and
  Levine}]{pmlr-v162-hansen-estruch22a}
Philippe Hansen-Estruch, Amy Zhang, Ashvin Nair, Patrick Yin, and Sergey
  Levine. 2022.
\newblock \href {https://proceedings.mlr.press/v162/hansen-estruch22a.html}
  {Bisimulation makes analogies in goal-conditioned reinforcement learning}.
\newblock In \emph{Proceedings of the 39th International Conference on Machine
  Learning}, volume 162 of \emph{Proceedings of Machine Learning Research},
  pages 8407--8426. PMLR.

\bibitem[{He et~al.(2023)He, Gao, and Chen}]{he2023debertav}
Pengcheng He, Jianfeng Gao, and Weizhu Chen. 2023.
\newblock \href {https://openreview.net/forum?id=sE7-XhLxHA} {De{BERT}av3:
  Improving de{BERT}a using {ELECTRA}-style pre-training with
  gradient-disentangled embedding sharing}.
\newblock In \emph{The Eleventh International Conference on Learning
  Representations}.

\bibitem[{Hesse(1959)}]{hesse1959defining}
Mary~B Hesse. 1959.
\newblock On defining analogy.
\newblock In \emph{Proceedings of the Aristotelian Society}, volume~60, pages
  79--100. JSTOR.

\bibitem[{Hofstadter(2001)}]{hofstadter2001analogy}
Douglas~R Hofstadter. 2001.
\newblock Analogy as the core of cognition.
\newblock \emph{The analogical mind: Perspectives from cognitive science},
  pages 499--538.

\bibitem[{Hofstadter and Sander(2013)}]{hofstadter2013surfaces}
Douglas~R Hofstadter and Emmanuel Sander. 2013.
\newblock \emph{Surfaces and essences: Analogy as the fuel and fire of
  thinking}.
\newblock Basic books.

\bibitem[{Hu et~al.(2023)Hu, Storks, Lewis, and Chai}]{hu-etal-2023-context}
Xiaoyang Hu, Shane Storks, Richard Lewis, and Joyce Chai. 2023.
\newblock \href {https://doi.org/10.18653/v1/2023.acl-long.109} {In-context
  analogical reasoning with pre-trained language models}.
\newblock In \emph{Proceedings of the 61st Annual Meeting of the Association
  for Computational Linguistics (Volume 1: Long Papers)}, pages 1953--1969,
  Toronto, Canada. Association for Computational Linguistics.

\bibitem[{Ilievski et~al.(2022)Ilievski, Pujara, and Shenoy}]{ilievski2022does}
Filip Ilievski, Jay Pujara, and Kartik Shenoy. 2022.
\newblock Does wikidata support analogical reasoning?
\newblock In \emph{Iberoamerican Knowledge Graphs and Semantic Web Conference},
  pages 178--191. Springer.

\bibitem[{Jiayang et~al.(2023)Jiayang, Qiu, Chan, Fang, Wang, Chan, Ru, Guo,
  Zhang, Song, Zhang, and Zhang}]{jiayang2023storyanalogy}
Cheng Jiayang, Lin Qiu, Tsz~Ho Chan, Tianqing Fang, Weiqi Wang, Chunkit Chan,
  Dongyu Ru, Qipeng Guo, Hongming Zhang, Yangqiu Song, Yue Zhang, and Zheng
  Zhang. 2023.
\newblock \href {http://arxiv.org/abs/2310.12874} {Storyanalogy: Deriving
  story-level analogies from large language models to unlock analogical
  understanding}.

\bibitem[{Levy and Goldberg(2014)}]{levy-goldberg-2014-linguistic}
Omer Levy and Yoav Goldberg. 2014.
\newblock \href {https://doi.org/10.3115/v1/W14-1618} {Linguistic regularities
  in sparse and explicit word representations}.
\newblock In \emph{Proceedings of the Eighteenth Conference on Computational
  Natural Language Learning}, pages 171--180, Ann Arbor, Michigan. Association
  for Computational Linguistics.

\bibitem[{Li et~al.(2020)Li, Yang, and Ma}]{li-etal-2020-ca}
Peng-Hsuan Li, Tsan-Yu Yang, and Wei-Yun Ma. 2020.
\newblock \href {https://aclanthology.org/2020.lrec-1.365} {{CA}-{EHN}:
  Commonsense analogy from {E}-{H}ow{N}et}.
\newblock In \emph{Proceedings of the Twelfth Language Resources and Evaluation
  Conference}, pages 2984--2990, Marseille, France. European Language Resources
  Association.

\bibitem[{Li et~al.(2018)Li, Zhao, Hu, Li, Liu, and
  Du}]{li-etal-2018-analogical}
Shen Li, Zhe Zhao, Renfen Hu, Wensi Li, Tao Liu, and Xiaoyong Du. 2018.
\newblock \href {https://doi.org/10.18653/v1/P18-2023} {Analogical reasoning on
  {C}hinese morphological and semantic relations}.
\newblock In \emph{Proceedings of the 56th Annual Meeting of the Association
  for Computational Linguistics (Volume 2: Short Papers)}, pages 138--143,
  Melbourne, Australia. Association for Computational Linguistics.

\bibitem[{Liu et~al.(2019)Liu, Ott, Goyal, Du, Joshi, Chen, Levy, Lewis,
  Zettlemoyer, and Stoyanov}]{liu2019roberta}
Yinhan Liu, Myle Ott, Naman Goyal, Jingfei Du, Mandar Joshi, Danqi Chen, Omer
  Levy, Mike Lewis, Luke Zettlemoyer, and Veselin Stoyanov. 2019.
\newblock Roberta: A robustly optimized bert pretraining approach.
\newblock \emph{arXiv preprint arXiv:1907.11692}.

\bibitem[{Martinez-Rodriguez et~al.(2018)Martinez-Rodriguez, Lopez-Arevalo, and
  Rios-Alvarado}]{MARTINEZRODRIGUEZ2018339}
Jose~L. Martinez-Rodriguez, Ivan Lopez-Arevalo, and Ana~B. Rios-Alvarado. 2018.
\newblock \href {https://doi.org/https://doi.org/10.1016/j.eswa.2018.07.017}
  {Openie-based approach for knowledge graph construction from text}.
\newblock \emph{Expert Systems with Applications}, 113:339--355.

\bibitem[{Mikolov et~al.(2013{\natexlab{a}})Mikolov, Sutskever, Chen, Corrado,
  and Dean}]{10.5555/2999792.2999959}
Tomas Mikolov, Ilya Sutskever, Kai Chen, Greg Corrado, and Jeffrey Dean.
  2013{\natexlab{a}}.
\newblock Distributed representations of words and phrases and their
  compositionality.
\newblock In \emph{Proceedings of the 26th International Conference on Neural
  Information Processing Systems - Volume 2}, NIPS'13, page 3111–3119, Red
  Hook, NY, USA. Curran Associates Inc.

\bibitem[{Mikolov et~al.(2013{\natexlab{b}})Mikolov, Yih, and
  Zweig}]{mikolov-etal-2013-linguistic}
Tomas Mikolov, Wen-tau Yih, and Geoffrey Zweig. 2013{\natexlab{b}}.
\newblock \href {https://aclanthology.org/N13-1090} {Linguistic regularities in
  continuous space word representations}.
\newblock In \emph{Proceedings of the 2013 Conference of the North {A}merican
  Chapter of the Association for Computational Linguistics: Human Language
  Technologies}, pages 746--751, Atlanta, Georgia. Association for
  Computational Linguistics.

\bibitem[{Miller(1995)}]{10.1145/219717.219748}
George~A. Miller. 1995.
\newblock \href {https://doi.org/10.1145/219717.219748} {Wordnet: A lexical
  database for english}.
\newblock \emph{Commun. ACM}, 38(11):39–41.

\bibitem[{Mitchell(2021)}]{mitchell_abstraction_2021}
Melanie Mitchell. 2021.
\newblock \href {https://doi.org/https://doi.org/10.1111/nyas.14619}
  {Abstraction and analogy-making in artificial intelligence}.
\newblock \emph{Annals of the New York Academy of Sciences}, 1505(1):79--101.

\bibitem[{OpenAI(2022)}]{openai2022chatgpt}
OpenAI. 2022.
\newblock \href {https://openai.com/blog/chatgpt} {Chatgpt}.

\bibitem[{Ouyang et~al.(2022)Ouyang, Wu, Jiang, Almeida, Wainwright, Mishkin,
  Zhang, Agarwal, Slama, Ray et~al.}]{ouyang2022training}
Long Ouyang, Jeff Wu, Xu~Jiang, Diogo Almeida, Carroll~L Wainwright, Pamela
  Mishkin, Chong Zhang, Sandhini Agarwal, Katarina Slama, Alex Ray, et~al.
  2022.
\newblock Training language models to follow instructions with human feedback.
\newblock \emph{arXiv preprint arXiv:2203.02155}.

\bibitem[{Raffel et~al.(2022)Raffel, Shazeer, Roberts, Lee, Narang, Matena,
  Zhou, Li, and Liu}]{10.5555/3455716.3455856}
Colin Raffel, Noam Shazeer, Adam Roberts, Katherine Lee, Sharan Narang, Michael
  Matena, Yanqi Zhou, Wei Li, and Peter~J. Liu. 2022.
\newblock Exploring the limits of transfer learning with a unified text-to-text
  transformer.
\newblock \emph{J. Mach. Learn. Res.}, 21(1).

\bibitem[{Reimers and Gurevych(2019)}]{reimers-gurevych-2019-sentence}
Nils Reimers and Iryna Gurevych. 2019.
\newblock \href {https://doi.org/10.18653/v1/D19-1410} {Sentence-{BERT}:
  Sentence embeddings using {S}iamese {BERT}-networks}.
\newblock In \emph{Proceedings of the 2019 Conference on Empirical Methods in
  Natural Language Processing and the 9th International Joint Conference on
  Natural Language Processing (EMNLP-IJCNLP)}, pages 3982--3992, Hong Kong,
  China. Association for Computational Linguistics.

\bibitem[{Romero and Razniewski(2020)}]{10.1145/3340531.3417416}
Julien Romero and Simon Razniewski. 2020.
\newblock \href {https://doi.org/10.1145/3340531.3417416} {Inside quasimodo:
  Exploring construction and usage of commonsense knowledge}.
\newblock In \emph{Proceedings of the 29th ACM International Conference on
  Information \&amp; Knowledge Management}, CIKM '20, page 3445–3448, New
  York, NY, USA. Association for Computing Machinery.

\bibitem[{Schluter(2018)}]{schluter-2018-word}
Natalie Schluter. 2018.
\newblock \href {https://doi.org/10.18653/v1/N18-2039} {The word analogy
  testing caveat}.
\newblock In \emph{Proceedings of the 2018 Conference of the North {A}merican
  Chapter of the Association for Computational Linguistics: Human Language
  Technologies, Volume 2 (Short Papers)}, pages 242--246, New Orleans,
  Louisiana. Association for Computational Linguistics.

\bibitem[{Si and Carlson(2017)}]{DBLP:conf/cogsci/SiC17}
Mei Si and Craig Carlson. 2017.
\newblock \href {https://mindmodeling.org/cogsci2017/papers/0595/index.html} {A
  data-driven approach for making analogies}.
\newblock In \emph{Proceedings of the 39th Annual Meeting of the Cognitive
  Science Society, CogSci 2017, London, UK, 16-29 July 2017}.

\bibitem[{Speer et~al.(2017)Speer, Chin, and Havasi}]{10.5555/3298023.3298212}
Robyn Speer, Joshua Chin, and Catherine Havasi. 2017.
\newblock Conceptnet 5.5: An open multilingual graph of general knowledge.
\newblock In \emph{Proceedings of the Thirty-First AAAI Conference on
  Artificial Intelligence}, AAAI'17, page 4444–4451. AAAI Press.

\bibitem[{Speer et~al.(2008)Speer, Havasi, and
  Lieberman}]{Speer2008AnalogySpaceRT}
Robyn Speer, Catherine Havasi, and Henry Lieberman. 2008.
\newblock \href {https://api.semanticscholar.org/CorpusID:8655598}
  {Analogyspace: Reducing the dimensionality of common sense knowledge}.
\newblock In \emph{AAAI Conference on Artificial Intelligence}.

\bibitem[{Turney et~al.(2003)Turney, Littman, Bigham, and
  Shnayder}]{turney2003combining}
Peter~D Turney, Michael~L Littman, Jeffrey Bigham, and Victor Shnayder. 2003.
\newblock Combining independent modules in lexical multiple-choice problems.
\newblock \emph{Recent Advances in Natural Language Processing III: Selected
  Papers from RANLP}, 2003:101--110.

\bibitem[{Ul{\v{c}}ar et~al.(2020)Ul{\v{c}}ar, Vaik, Lindstr{\"o}m,
  Dailid{\.e}nait{\.e}, and
  Robnik-{\v{S}}ikonja}]{ulcar-etal-2020-multilingual}
Matej Ul{\v{c}}ar, Kristiina Vaik, Jessica Lindstr{\"o}m, Milda
  Dailid{\.e}nait{\.e}, and Marko Robnik-{\v{S}}ikonja. 2020.
\newblock \href {https://www.aclweb.org/anthology/2020.lrec-1.501}
  {Multilingual culture-independent word analogy datasets}.
\newblock In \emph{Proceedings of the 12th Language Resources and Evaluation
  Conference}, pages 4074--4080, Marseille, France. European Language Resources
  Association.

\bibitem[{Ushio et~al.(2021)Ushio, Espinosa~Anke, Schockaert, and
  Camacho-Collados}]{ushio-etal-2021-bert}
Asahi Ushio, Luis Espinosa~Anke, Steven Schockaert, and Jose Camacho-Collados.
  2021.
\newblock \href {https://doi.org/10.18653/v1/2021.acl-long.280} {{BERT} is to
  {NLP} what {A}lex{N}et is to {CV}: Can pre-trained language models identify
  analogies?}
\newblock In \emph{Proceedings of the 59th Annual Meeting of the Association
  for Computational Linguistics and the 11th International Joint Conference on
  Natural Language Processing (Volume 1: Long Papers)}, pages 3609--3624,
  Online. Association for Computational Linguistics.

\bibitem[{Vrande\v{c}i\'{c} and Kr\"{o}tzsch(2014)}]{10.1145/2629489}
Denny Vrande\v{c}i\'{c} and Markus Kr\"{o}tzsch. 2014.
\newblock \href {https://doi.org/10.1145/2629489} {Wikidata: A free
  collaborative knowledgebase}.
\newblock \emph{Commun. ACM}, 57(10):78–85.

\bibitem[{Webb et~al.(2023)Webb, Holyoak, and Lu}]{webb2023emergent}
Taylor Webb, Keith~J Holyoak, and Hongjing Lu. 2023.
\newblock Emergent analogical reasoning in large language models.
\newblock \emph{Nature Human Behaviour}, pages 1--16.

\bibitem[{Wijesiriwardene et~al.(2023)Wijesiriwardene, Wickramarachchi, Gajera,
  Gowaikar, Gupta, Chadha, Reganti, Sheth, and
  Das}]{wijesiriwardene-etal-2023-analogical}
Thilini Wijesiriwardene, Ruwan Wickramarachchi, Bimal Gajera, Shreeyash
  Gowaikar, Chandan Gupta, Aman Chadha, Aishwarya~Naresh Reganti, Amit Sheth,
  and Amitava Das. 2023.
\newblock \href {https://doi.org/10.18653/v1/2023.findings-acl.218}
  {{ANALOGICAL} - a novel benchmark for long text analogy evaluation in large
  language models}.
\newblock In \emph{Findings of the Association for Computational Linguistics:
  ACL 2023}, pages 3534--3549, Toronto, Canada. Association for Computational
  Linguistics.

\bibitem[{Wu et~al.(2012)Wu, Li, Wang, and Zhu}]{10.1145/2213836.2213891}
Wentao Wu, Hongsong Li, Haixun Wang, and Kenny~Q. Zhu. 2012.
\newblock \href {https://doi.org/10.1145/2213836.2213891} {Probase: A
  probabilistic taxonomy for text understanding}.
\newblock In \emph{Proceedings of the 2012 ACM SIGMOD International Conference
  on Management of Data}, SIGMOD '12, page 481–492, New York, NY, USA.
  Association for Computing Machinery.

\bibitem[{Yamada et~al.(2020)Yamada, Asai, Shindo, Takeda, and
  Matsumoto}]{yamada-etal-2020-luke}
Ikuya Yamada, Akari Asai, Hiroyuki Shindo, Hideaki Takeda, and Yuji Matsumoto.
  2020.
\newblock \href {https://doi.org/10.18653/v1/2020.emnlp-main.523} {{LUKE}: Deep
  contextualized entity representations with entity-aware self-attention}.
\newblock In \emph{Proceedings of the 2020 Conference on Empirical Methods in
  Natural Language Processing (EMNLP)}, pages 6442--6454, Online. Association
  for Computational Linguistics.

\bibitem[{Yasunaga et~al.(2023)Yasunaga, Chen, Li, Pasupat, Leskovec, Liang,
  Chi, and Zhou}]{yasunaga2023large}
Michihiro Yasunaga, Xinyun Chen, Yujia Li, Panupong Pasupat, Jure Leskovec,
  Percy Liang, Ed~H Chi, and Denny Zhou. 2023.
\newblock Large language models as analogical reasoners.
\newblock \emph{arXiv preprint arXiv:2310.01714}.

\bibitem[{Zhang et~al.(2022)Zhang, Li, Chen, Liang, Deng, and
  Chen}]{zhang2022multimodal}
Ningyu Zhang, Lei Li, Xiang Chen, Xiaozhuan Liang, Shumin Deng, and Huajun
  Chen. 2022.
\newblock Multimodal analogical reasoning over knowledge graphs.
\newblock \emph{arXiv preprint arXiv:2210.00312}.

\bibitem[{Zhang et~al.(2019)Zhang, Han, Liu, Jiang, Sun, and
  Liu}]{zhang-etal-2019-ernie}
Zhengyan Zhang, Xu~Han, Zhiyuan Liu, Xin Jiang, Maosong Sun, and Qun Liu. 2019.
\newblock \href {https://doi.org/10.18653/v1/P19-1139} {{ERNIE}: Enhanced
  language representation with informative entities}.
\newblock In \emph{Proceedings of the 57th Annual Meeting of the Association
  for Computational Linguistics}, pages 1441--1451, Florence, Italy.
  Association for Computational Linguistics.

\end{thebibliography}
%\bibliographystyle{acl_natbib}

%\clearpage
\appendix

\section{Details of \method}
\subsection{Terminology Definition in \method}\label{appendix:define}

To better understand the schema for analogies in \method, we list the terminologies in Table~\ref{tab:terminology}.

\subsection{Crowd-sourcing Details}\label{appendix:crowdsourcing}
\begin{figure*}[t]
    \centering
    \small
    \includegraphics[width=\linewidth]{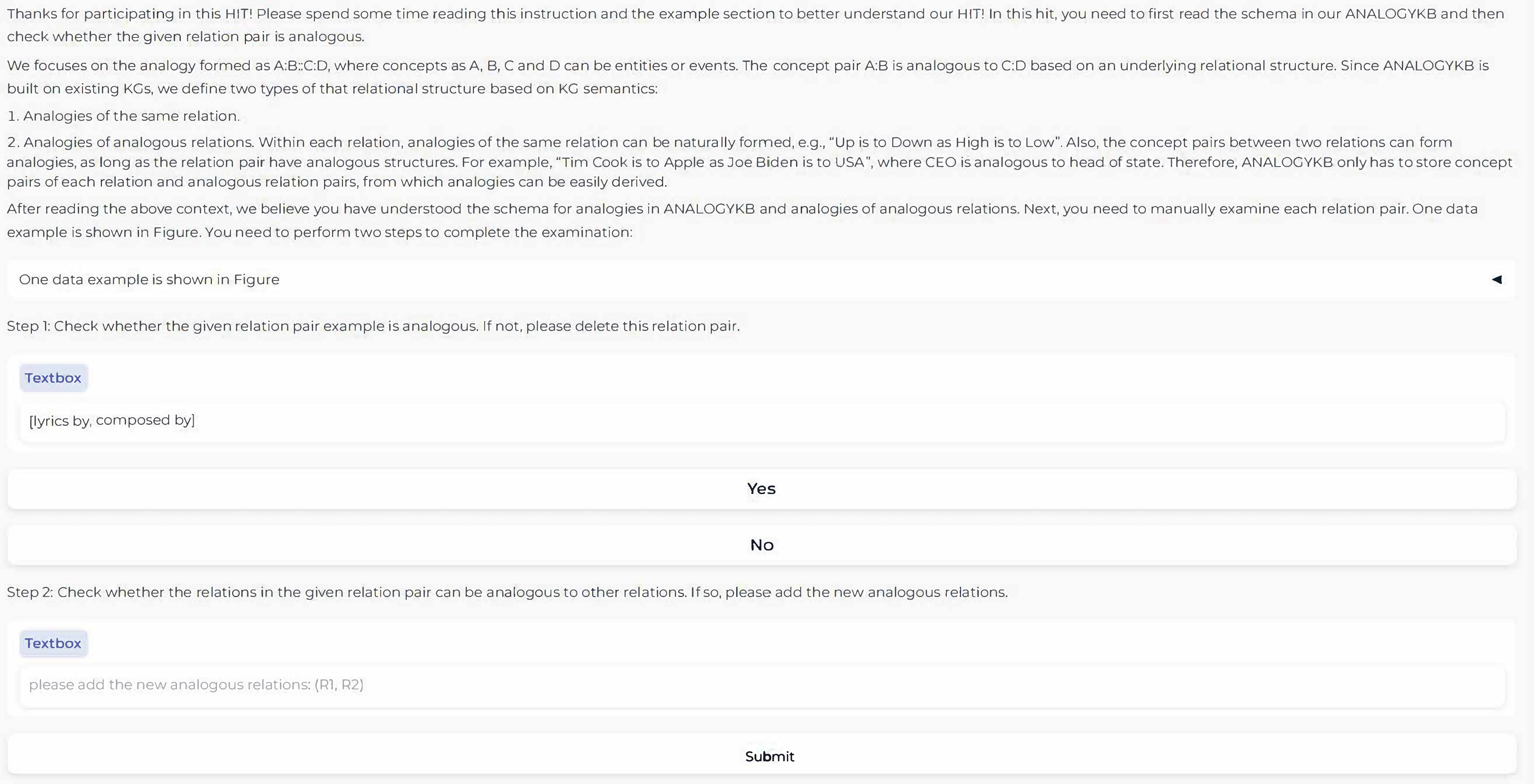}
    \caption{The screenshots of the instructions and annotation interface.
    }
    \label{fig:annotation}
\end{figure*}

We have recruited a team of two undergraduates. We pay each annotator \$8/h, exceeding the local minimum wage.
The screenshots of the instructions and annotation interface are shown in Figure~\ref{fig:annotation}.

\section{Benchmark}\label{appendix:Benchmark} 
We compare our methods with baselines and human performance in 6 different benchmarks.
An example of these benchmarks is given in Table~\ref{tab:example_word_analogy}
For benchmarks without training sets, we only fine-tune models on their validation sets.
\begin{itemize}
    \item \textbf{E-KAR}~\cite{chen-etal-2022-e}: a \textbf{E}xplainable \textbf{K}nowledge-intensive \textbf{A}nalogical \textbf{R}easoning benchmark sourced from the publicly available Civil Service Examinations (CSE) of China, which contains linguistic, commonsense, encyclopedic, and cultural (e.g., idiom and historical) knowledge. 
    This dataset contains 870 training data, 119 validation data, and 262 test data.
    The SOTA model on this benchmark is proposed by \citet{chen-etal-2022-e}.
    \item \textbf{BATS}~\cite{gladkova-etal-2016-analogy}: is \textbf{B}igger \textbf{A}nalogy \textbf{T}est \textbf{S}et containing more than 1,000 analogies.
    The analogies can be divided into four categories: lexicographic, encyclopedic, derivational and inflectional morphology. 
    This dataset contains 199 validation data and 1799 test data.
    The SOTA model on this benchmark is proposed by \citet{ushio-etal-2021-bert}.
    \item \textbf{UNIT 2}~\cite{Boteanu_Chernova_2015}: a benchmark using word analogy problems from an educational resource. 
    This dataset contains 24 validation data and 228 test data.
    The SOTA model on this benchmark is proposed by \citet{ushio-etal-2021-bert}.
    \item \textbf{UNIT 4}~\cite{Boteanu_Chernova_2015}: this benchmark also comes from an educational resource but is harder than U2. 
    This dataset contains 48 validation data and 432 test data.
    The SOTA model on this benchmark is proposed by \citet{ushio-etal-2021-bert}
    \item \textbf{Google}~\cite{mikolov-etal-2013-linguistic}: a benchmark for intrinsic evaluation of word embeddings proposed by Google, which contains semantic and morphological relations. 
    This dataset consists of 50 validation data and 500 test data.
    The SOTA model on this benchmark is proposed by \citet{chen-etal-2022-e}
    \item \textbf{SAT}~\cite{turney2003combining}: a benchmark constructed from SAT exams in the US college admission test consisting of 374 word analogy problems. 
    This dataset contains 37 validation data and 337 test data.
    The SOTA model on this benchmark is proposed by \citet{ushio-etal-2021-bert}.
\end{itemize}

As shown in Table~\ref{tab:overlap}, We list the overlap rates of \method with other analogy datasets. 
The overlap rates are calculated as (Data in \method
Data in Other Datasets) / (Data in Other Datasets). 
Specifically, one data sample, i.e., "A is to B as C is to D" can be changed into two tuples (A, R1, B) and (C, R2, D), where R1 and R2 can be exactly the same or analogous. 
If both tuples are present in \method, the overlap rate for this data instance is considered greater than 0.
The results indicate that \method contains a portion of the data from other analogy benchmarks, exhibiting high coverage. 
However, after our checking, we confirm that the training data sampled from \method, which is used to train LMs, does not contain the test data from other analogy benchmarks. 
This confirms the absence of data leakage, underscoring that LMs on \method can significantly improve the model performance on analogy recognition and generation tasks.

\begin{table}[t]
\small
  \centering
    \begin{tabular}{ll}
    \toprule
    Query & army:order \\
    \cdashlinelr{1-2}
    Candidates: & (A) volunteer:summon\\
                & \color[rgb]{0,0.39,0}\textit{\textbf{(B) band:band leader}}\\
                & (C) tourist:guide \\
                & (D) students:instruction\\
    \bottomrule
    \end{tabular}
  \caption{An example of analogy recognition task.
  The true answers are {\color[rgb]{0,0.39,0}\textit{\textbf{highlighted}}}.
  }
  \label{tab:example_word_analogy}
\end{table}

\begin{table*}[t]
  \centering
  \small
    \begin{tabular}{lll}
     \toprule
\textbf{Category} & \textbf{Definition} & \textbf{Example} \\ 
\midrule
Analogies & A:B::C:D (A is to B as C is to D) & \makecell[l]{Up:Down::High:Low, \\Tim Cook:Apple::Joe Biden:USA} \\
\midrule
Concept pairs & A:B or C:D & \makecell[l]{Left:Right, \\Tim Cook:Apple} \\ 
\midrule
Relation pairs & Two relations & \makecell[l]{(antonym, CEO), \\(CEO, head of state)} \\ 
\midrule
Analogous relations & Two relations that can form analogies & (CEO, head of state) \\
\midrule
Analogies of analogous relations & \makecell[l]{A:B::C:D where the relation of A:B is \\different but analogous to the relation of C:D} & Tim Cook:Apple::Joe Biden:USA \\ 
\bottomrule
    \end{tabular}%
    \caption{The definitions of terminologies with examples in the schema for \method}
  \label{tab:terminology}%
\end{table*}%

\begin{table}[t]
  \centering
  \small
   \begin{tabular}{lc}
    \toprule
    \textbf{Dataset}&	\textbf{Overlap Rate} \\
    \midrule
    E-KAR&	28.98\%\\
    BATS&	78.25\%\\
    UNIT 2&	52.32\%\\
    UNIT 4&	41.48\%\\
    Google&	98.52\%\\
    SAT&	34.70\%\\
    \bottomrule
    \end{tabular}%
    \caption{The overlap rates of \method with other analogy datasets.}
  \label{tab:overlap}%
\end{table}%

\section{Analogy Recognition Task}\label{appendix:ART}
\subsection{Data Construction}\label{appendix:ART_dataset}
To pre-train RoBERTa-Large on \method, we randomly sample 5,000 analogies of the same relation and 5,000 analogies of analogous relations from \method and formulate them into the multiple-choice question-answering format.
Specifically, for each instance, we randomly sample a concept pair from \method as a query and select another concept pair from the analogous relation as the answer.
Then, we randomly sample 3 concept pairs from the relations that can not be analogous to the relation of the query as distractions.

We also randomly sample 10,000 data of the same relations as an ablated variant to show the effectiveness of analogies of analogous relations (denoted as \textbf{Data$_\texttt{same}$}).
The construction method is similar, except that the query and answer are derived from the same relation.
Additionally, we randomly sample 10,000 data points from \method and construct analogy-style data (denoted as \textbf{Data$_\texttt{pseudo}$}).
Specifically, we randomly sample 50,000 concept pairs without considering analogous relations from \method as the data pool.
For each data point, we randomly sample 5 concept pairs from the data pool and choose one as the query, one as the answer, and the remaining three as distractions.

\subsection{Details of Baselines}\label{appendix:models}
\paragraph{Word Embedding and Sentence Embedding}
For the method of pre-trained word embeddings, we follow the method proposed by \citet{ushio-etal-2021-bert}.
And represent word pairs by taking the difference between their embeddings. Then, we choose the answer candidate with the highest cosine similarity to the query in terms of this vector difference.
For the method of sentence embedding, we convert query A:B to "A is to B" and choose the answer candidate ("C is to D") with the highest cosine similarity to the query.

\subsection{Training Process}\label{appendix:ART_training}
To pre-train language models on the sample data from \method, we follow the code from Huggingface~\footnote{\url{https://huggingface.co/docs/transformers/tasks/multiple_choice}}.
Since previous benchmarks, except E-KAR, do not have a training set, we fine-tune LMs on their small development set (about 300 samples). 
To achieve hyperparameter search, we maximize performance on the development set of E-KAR (119 data samples) as a compromise.
The training settings are: batch size = 64, learning rate = 3e-5, dropout rate = 0.1 and training epoch = 10.

\begin{figure}[t]
    \centering
    \small
    \includegraphics[width=\linewidth]{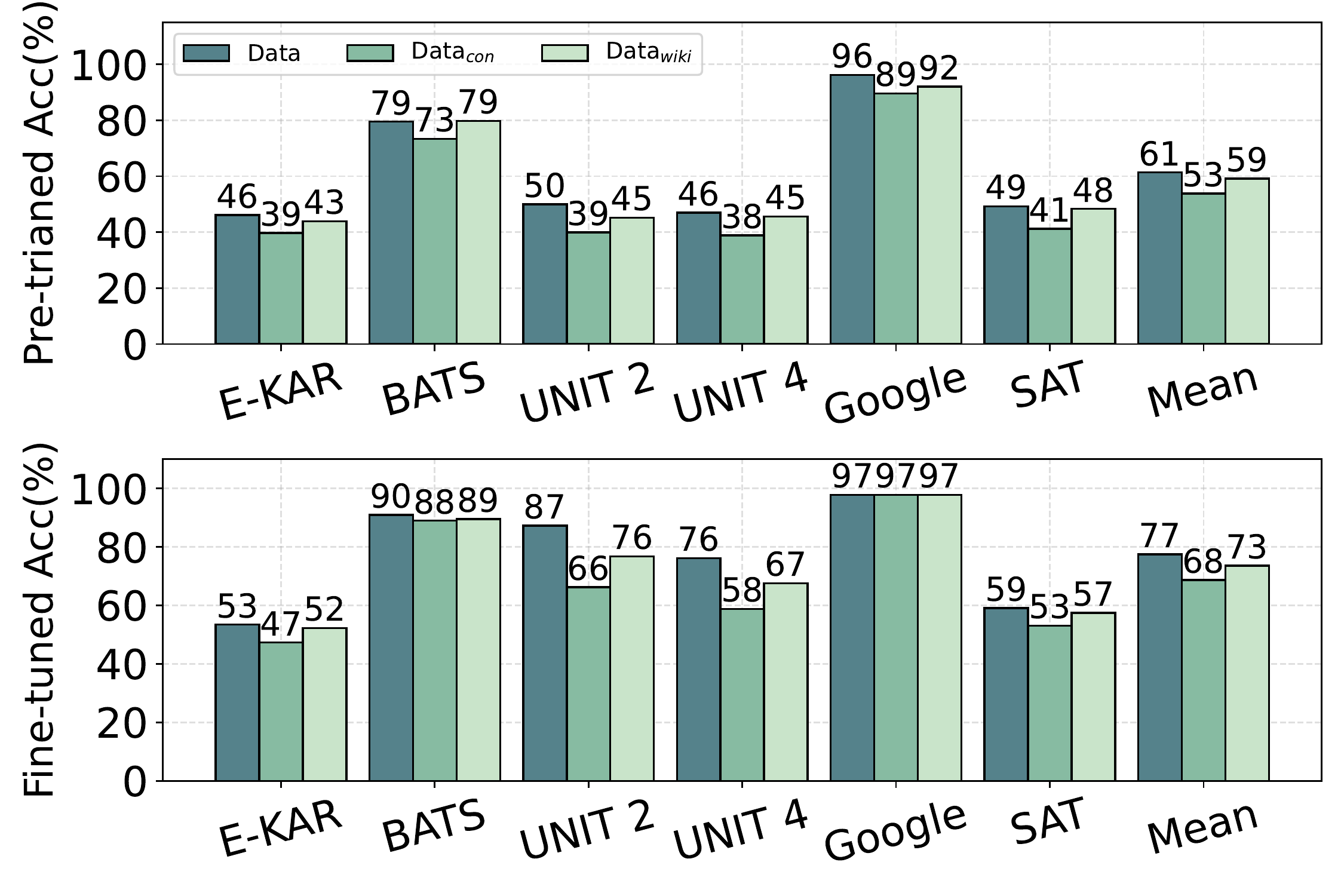}
    \caption{The accuracy of RoBERTa-Large trained on different data subsets on the analogy recognition task. 
    \textbf{Data} denotes the dataset sampled directly from \method and \textbf{Data$_\texttt{con}$} (or \textbf{Data$_\texttt{wiki}$}) denotes the analogies only from \conceptnet (or \wikidata).
    All the datasets have the same size.
    }
    \label{fig:ablation_KB}
\end{figure}
\begin{table}[t]
\footnotesize
  \centering
    \begin{tabular}{l}
    \toprule
    %\rowcolor[gray]{0.95}\multicolumn{1}{c}{\textbf{Analogy Retrieval Task}}
    %\midrule
    \makecell[l]{
    \color{gray}{/* \textit{Task prompt} */}\\
    Please make analogies.\\
    \color{gray}{/* \textit{Examples} */} \\
    input: artist is to paintbrush as magician is to\\
    output: wand\\
    input: razor is to shave as knife is to \\
    output: cut \\
    ...\\
    \color{gray}{/* \textit{Test data} */} \\
    input: classroom is to desk as church is to\\
    output: \color[rgb]{0,0.39,0}\textit{pew}
    } \\
    \bottomrule
    \end{tabular}
  \caption{Prompt for LLMs for analogy generation task. Generated texts by LLMs are {\color[rgb]{0,0.39,0}\textit{highlighted}}.
  }
  \label{tab:GPT-3_prompt_for_retrieve}
\end{table}

\setlength\tabcolsep{4pt}
%\setlength\tabrowsep{4pt}
\begin{table}[t]
  \centering
  \small
    \begin{tabular}{lcccc}
    \toprule
    \textbf{Data Size} & \textbf{Hit@$k$} & \textbf{E-KAR} & \textbf{UNIT 4} & \textbf{SAT} \\
    \midrule
    \multirow{3}[0]{*}{100K} & 1 & 30.00 & 38.00 & 25.00 \\
                & 3 & 33.00 & 44.00 & 25.00 \\
                & 5 & 33.00 & 44.00 & 26.00 \\
                 \cdashlinelr{1-5}
    \multirow{3}[0]{*}{500K} & 1 & 39.00 & 53.00 & 38.00 \\
                 & 3 & 42.00 & 58.00 & 38.00 \\
                 & 5 & 42.00 & 63.00 & 41.00 \\
                  \cdashlinelr{1-5}
    \multirow{3}[0]{*}{1M}   & 1 & 57.00 & 80.00 & 64.00 \\
                 & 3 & 62.00 & 86.00 & 76.00 \\
                 & 5 & \textbf{66.00} & \textbf{91.00} & \textbf{84.00} \\
    \bottomrule
    \end{tabular}
    \caption{The model trained on data with different sizes is T5-Large (770M).
    %We only obtain one generation with the highest probability (\textbf{Hit@1}) from InstructGPT through OpenAI API.
    }
  \label{tab:gen_datasize}%
\end{table}%

\setlength\tabcolsep{4pt}
%\setlength\tabrowsep{4pt}
\begin{table}[t]
  \centering
  \small
    \begin{tabular}{lcccc}
    \toprule
    \textbf{Model Size} & \textbf{Hit@$k$} & \textbf{E-KAR} & \textbf{UNIT 4} & \textbf{SAT} \\
    \midrule
    \multirow{3}[0]{*}{T5-small (60M)} & 1 & 18.00 & 18.00 & 14.00 \\
     & 3 & 18.00 & 21.00 & 15.00 \\
     & 5 & 18.00 & 22.00 & 15.00 \\
      \cdashlinelr{1-5}
    \multirow{3}[0]{*}{T5-base (220M)} & 1 & 22.00 & 31.00 & 28.00 \\
     & 3 & 23.00 & 31.00 & 34.00 \\
     & 5 & 25.00 & 31.00 & 34.00 \\
      \cdashlinelr{1-5}
    \multirow{3}[0]{*}{T5-large (770M)} & 1 & 57.00 & 80.00 & 64.00 \\
     & 3 & 62.00 & 86.00 & 76.00 \\
     & 5 & \textbf{66.00} & \textbf{91.00} & \textbf{84.00} \\
    \bottomrule
    \end{tabular}
    \caption{The model trained on data with different sizes is T5-Large (770M).
    %We only obtain one generation with the highest probability (\textbf{Hit@1}) from InstructGPT through OpenAI API.
    }
  \label{tab:gen_modelsize}%
\end{table}%

\begin{table*}[t]
  \centering
  \small
    \begin{tabular}{lcccccc}
    \toprule

\textbf{Model} & \textbf{E-KAR} & \textbf{UNIT 4} & \textbf{SAT} & \textbf{BATS} & \textbf{UNIT 2} & \textbf{Google} \\
\midrule
vanilla T5 & 13.00 & 17.00 & 8.00 & 38.00 & 35.00 & 45.00 \\
AnalogyT5$_\texttt{same}$ & 42.00 & 63.00 & 37.00 & 75.00 & 73.00 & 94.00 \\
AnalogyT5 & \textbf{57.00} & \textbf{80.00} & \textbf{64.00} & \textbf{80.00} & \textbf{84.00} & \textbf{95.00} \\
\midrule
InstructGPT$_\texttt{003}$ & 61.00 & 70.00 & 60.00 & 82.00 & 79.00 & 94.00 \\
+ Human & 68.00 & 76.00 & 74.00 & 85.00 & 83.00 & 98.00 \\
+ \textsc{AnalogyKB}$_\texttt{same}$ & 64.00 & 77.00 & 77.00 & 83.00 & 85.00 & \textbf{100.00} \\
+ \textsc{AnalogyKB} & \textbf{75.00} & 80.00 & 85.00 & 88.00 & 88.00 & \textbf{100.00} \\
\cdashlinelr{1-7}
ChatGPT & 58.00 & 76.00 & 78.00 & 84.00 & 84.00 & 96.00 \\
+ Human & 64.00 & 81.00 & 80.00 & 88.00 & 88.00 & \textbf{100.00} \\
+ \textsc{AnalogyKB}$_\texttt{same}$ & 64.00 & 80.00 & 81.00 & 92.00 & 91.00 & \textbf{100.00} \\
+ \textsc{AnalogyKB} & 69.00 & \textbf{92.00} & \textbf{91.00} & \textbf{96.00} & \textbf{94.00} & \textbf{100.00} \\
\bottomrule
\end{tabular}
\caption{
    The accuracy of different methods on the six analogy benchmark tasks in the analogy generation task.
    }
  \label{tab:retrieve_all}%
\end{table*}%
\begin{table}[t]
  \centering
  \small
    \begin{tabular}{ll}
    \toprule
    \textbf{Input}  & \textbf{Completion} \\
    \midrule
    \textit{Mcdonald} is to \textit{America} as \textit{Samsung} is to & \green{south korea} \\
    \textit{oxygen} is to \textit{breathe} as \textit{brain} is to & \green{thinking} \\
    \textit{terrestrial} is to \textit{land} as \textit{aquatic} is to & \green{water} \\
    \textit{meticulous} is to \textit{careful} as \textit{ascetic} is to  & \red{asceticism} \\
    \textit{triangle} is to \textit{area} as \textit{cube} is to & \green{volume} \\
    \cdashlinelr{1-2}
    \textit{electron} is to \textit{nucleus} as \textit{earth} is to & \green{sun} \\
    \textit{electron} is to \textit{electric force} as \textit{earth} is to & \green{gravity} \\
    \textit{electron} is to \textit{atom} as \textit{earth} is to & \green{solar system} \\
    \bottomrule
    \end{tabular}%
    \caption{Randomly selected and novel analogy generated from the AnalogyT5. 
    Novel generations are concept pairs not found in the training set of AnalogyT5.
    Whether the analogy is considered \green{plausible} or \red{not} is decided by human annotators.}
  \label{tab:case}%
\end{table}%

% Table generated by Excel2LaTeX from sheet 'Sheet2'
\begin{table}[t]
  \centering
  \small
    \begin{tabular}{ccll}
    \toprule
    \multicolumn{1}{l}{\textbf{Target}} & \multicolumn{1}{l}{\textbf{Source}} & \textbf{Attribute} & \textbf{mapping} \\
    \midrule
    \multirow{5}[0]{*}{Argument} & \multirow{5}[0]{*}{War} & Debater & Combatant \\
          &       & Topic & Battleground \\
          &       & Claim & Position \\
          &       & Criticize & Attack \\
          &       & Rhetoric & Maneuver \\
    \bottomrule
    \end{tabular}%
    \caption{Example mappings in SCAN. For a source concept, multiple related attributes are mapped to corresponding attributes of the target concept.}
  \label{tab:scan_example}%
\end{table}%

\subsection{Comparison with Different KB Sources}\label{appendix:different_KB}
We also create two ablated variants to train the models to evaluate the necessity of \conceptnet and \wikidata:
\begin{inparaenum}[\it 1)]
    \item \textit{Analogies from \conceptnet}, denoted as \textbf{Data$_\texttt{con}$}: we randomly sampled 10,000 (the same size as before) data of the relations only in \conceptnet as an ablated variant.
    \item \textit{Analogies from \wikidata}, denoted as \textbf{Data$_\texttt{wiki}$}: we randomly sampled 10,000 data of the relations only in \wikidata as an ablated variant.
\end{inparaenum}
The results in Figure~\ref{fig:ablation_KB} show that \method can combine the commonsense knowledge of \conceptnet and the entity knowledge of \wikidata and thus exhibits superior performance in improving the analogy-making ability of models compared to utilizing a single data source. 

\subsection{Significant Test}\label{sec:significant}
For the results in Table~\ref{tab:evaluation}, we demonstrate that the random sampling of data does not greatly impact the accuracy through the statistical significance test.
Specifically, we sample the training data of \method twice with different random seeds and run our method on these benchmarks in Table~\ref{tab:evaluation}. 
Then, we implement a t-test on the two results with a 0.05 significance level. 
The result is not significant (p-value: 0.208), and thus we can not reject the null hypothesis (H0: $r_{1}-r_{2}=0$, where $r_{i}$=(Acc. of E-KAR, Acc. of BATS, Acc. of UNIT 2, Acc. of UNIT 4, Acc. of Google, Acc. of SAT));
Furthermore, we fix the training data of \method and run our method on the benchmarks in Table~\ref{tab:evaluation} twice with different random seeds. The result is insignificant (p-value: 0.250), and thus, we can not reject the null hypothesis (H0: $r_{1}-r_{2}=0$).

For the results in Figure~\ref{fig:ablation}, we conduct a statistical significance test on Data and Data$_\texttt{same}$.
We average the accuracy of the two settings and implement a t-test with a 0.05 significance level.
The null hypothesis H0 is $r_{1}-r_{2}=0$, and the H1 is $r_{1}-r_{2}>0$, where $r_{1}$ and $r_{2}$ are the lists of benchmarks' average accuracy of Data and Data$_\texttt{same}$ in Pre-trained and Fine-tuned settings.
The result is significant (p-value: 0.012), and we can reject the null hypothesis H0. 
Thus, we can conclude that analogies of analogous relations in \method are rather important for models in the analogy recognition task.

\section{Analogy Generation Task}\label{appendix:Retrieval}
\subsection{Training Process}\label{appendix:Retrieval_training}
To construct the training data, we convert A:B::C:D to ``\textit{A is to B as C is to D}'' and let T5-Large generate the concept D given the input text ``\textit{A is to B as C is to}''.
The training settings are: batch size = 32, learning rate = 3e-5, dropout rate = 0.1 and training epoch = 20.

\subsection{The impact of data sizes and model sizes}\label{sec:impact}

For the analogy generation task, we have examined the effects of training data size and model size on model performance. The results in Fugure~\ref{tab:gen_datasize} and Fugure~\ref{tab:gen_modelsize} show that:
\begin{inparaenum}[\it 1)]
    \item By incorporating a larger volume of data from \method, we observe a gradual improvement in model performance, revealing the essential role of \method.
    \item Only larger models with enough training data can boost the ability to generate reasonable analogies.
\end{inparaenum}

\subsection{Results on Six Benchmarks in Analogy Generation Tasks}\label{sec:six_benchmarks}
We expanded the experiments in Table~\ref{tab:retrieve} to six analogy benchmark tasks. 
The results in Table~\ref{tab:retrieve_all} indicate that
compared to analogies with simple and same relations, \method is more crucial for models to understand analogies with more abstract and complex relations, such as E-KAR, UNIT 4, and SAT.

\subsection{Case Study}\label{sec:case_study}
We are curious whether LMs trained on \method can generalize to novel analogies.
After manual inspection, we observe from Table~\ref{tab:case} that, AnalogyT5 can generate a reasonable concept D for the input.
AnalogyT5 also generates reasonable analogies of analogous relations, such as ``\textit{triangle}'' is to ``\textit{area}'' as ``\textit{cube}'' is to ``\textit{volume}''.
However, analogies about adjectives are more error-prone, possibly due to the paucity of adjectives in \method.
We also discover that training on \method enables LMs to generate reasonable analogies by changing concept B while holding fixed A (\ie, \textit{electron}) and C (\ie, \textit{earth}).

\subsection{Out-of-domain Analogy}\label{appendix:DAT}
\paragraph{Dataset}
SCAN~\cite{czinczoll2022scientific} is an analogy dataset consisting of 449 analogy instances clustered into 65 full-concept mappings.
The overlap rate of \method with SCAN is only 2.67\%.
An example mapping in SCAN is shown in Table~\ref{tab:scan_example}.
Unlike the previous analogy dataset, SCAN mainly contains metaphorical and scientific analogies, which are abstract and thus rarely appear in the corpus and are difficult for LMs.
In addition, each concept in SCAN only has one token and SCAN is not confined to the word analogy task due to its full-concept mappings.

\paragraph{Baseline}
The original paper evaluates the analogical capabilities of GPT-2 and BERT on the SCAN dataset.
The authors convert the analogy instance to ``\textit{If A is like B, then C is like D}'', and force the models to predict the last token of the sentence.
For GPT-2, the model needs to generate the last token given the input text ``\textit{If A is like B, then C is like}''.
For BERT, the authors first mask D as ``\textit{If A is like B, then C is like \mask}'' and let the model predict word D.

In addition, the authors fine-tune the LMs on the 1,500-sized set of BATS (\ie, + BATS) and investigate whether the models learn about analogical reasoning in general after training on BATS. 
We follow this setting and randomly sample 1,500 data from \method and fine-tune the LMs on the sample data (\ie, + \method).
To prove the necessity of analogies of analogous relations, we randomly sample 1,500 analogies of the same relations as an ablated variant (\ie, + \textsc{\method}$_\texttt{same}$).
We also added LMs trained on the 800 data points of E-KAR (\ie, + E-KAR).

We further explore the performance of LLMs on the SCAN dataset.
Specifically, we also adopt the prompt in Table~\ref{tab:GPT-3_prompt_for_retrieve} to let LLMs generate the word D.
Since each concept in SCAN has only one token, we can obtain the top 5 results from InstructGPT through the OpenAI API.

\paragraph{Evaluation Metrics}
Following \citet{czinczoll2022scientific}, we report accuracy, recall@5 and the mean reciprocal rank (MRR) to compare the performance of models.
To reduce computing, we only consider the MRR of the first token of the target word among the top 10 predicted tokens.
The RR of a label is 0 if it is not in the top 10 tokens.
\paragraph{Training Process}
The training settings of GPT-2 and BERT are: batch size = 128, learning rate = 3e-5, dropout rate = 0.1 and training epoch = 10.

\label{sec:appendix}
\end{document}